\documentclass[letterpaper]{article} 
\usepackage{aaai2026}  
\usepackage{times}  
\usepackage{helvet}  
\usepackage{courier}  
\usepackage[hyphens]{url}  
\usepackage{graphicx} 
\usepackage{natbib}  
\usepackage{caption} 
\usepackage{algorithm}
\usepackage{algorithmicx}
\usepackage{algpseudocode}
\usepackage{booktabs}
\usepackage{multirow}
\usepackage{amsmath}
\usepackage{amsfonts}

\usepackage{newfloat}
\usepackage{listings}
\DeclareCaptionStyle{ruled}{labelfont=normalfont,labelsep=colon,strut=off} 
\floatstyle{ruled}
\newfloat{listing}{tb}{lst}{}
\floatname{listing}{Listing}
\title{SemanticNN: Compressive and Error-Resilient Semantic Offloading for Extremely Weak Devices}
\author{
    Jiaming Huang, Yi Gao\thanks{Corresponding authors.}, Fuchang Pan, Renjie Li, and Wei Dong$^*$\\
}
\affiliations{
    Zhejiang Key Lab. of Accessible Perception and Intelligent Systems\\
    College of Computer Science, Zhejiang University, China\\

    \{huangjm, gaoyi, panfc, lirj, dongw\}@zju.edu.cn
}

\usepackage{bibentry}

\begin{document}

\maketitle

\begin{abstract}
With the rapid growth of the Internet of Things (IoT), integrating artificial intelligence (AI) on extremely weak embedded devices has garnered significant attention, enabling improved real-time performance and enhanced data privacy.
However, the resource limitations of such devices and unreliable network conditions necessitate error-resilient device-edge collaboration systems.
Traditional approaches focus on bit-level transmission correctness, which can be inefficient under dynamic channel conditions. 
In contrast, we propose SemanticNN, a semantic codec that tolerates bit-level errors in pursuit of semantic-level correctness, enabling compressive and resilient collaborative inference offloading under strict computational and communication constraints.
It incorporates a Bit Error Rate (BER)-aware decoder that adapts to dynamic channel conditions and a Soft Quantization (SQ)-based encoder to learn compact representations.
Building on this architecture, we introduce Feature-augmentation Learning, a novel training strategy that enhances offloading efficiency.
To address encoder-decoder capability mismatches from asymmetric resources, we propose XAI-based Asymmetry Compensation to enhance decoding semantic fidelity.
We conduct extensive experiments on STM32 using three models and six datasets across image classification and object detection tasks.
Experimental results demonstrate that, under varying transmission error rates, SemanticNN significantly reduces feature transmission volume by 56.82–344.83× while maintaining superior inference accuracy.
\end{abstract}

\begin{links}
    \link{Code}{https://github.com/zju-emnets/SemanticNN}
\end{links}

\section{Introduction}

Internet of Things (IoT) has become increasingly prevalent across a wide range of areas, including manufacturing, agriculture, transportation, and healthcare~\cite{quy2022iot,breda2023feverphone,ling2021rt, jiang2021flexible}.
Deploying intelligent capabilities on weak embedded devices in these settings enhances real-time responsiveness and operational efficiency~\cite{yao2020deep, huang2022real,maj2022deep, benazir2024speech}.
For instance, deploying neural networks (NNs) on small drones equipped with IoT cameras enables real-time anomaly detection, which benefits applications such as fire detection, wildlife monitoring, and traffic surveillance~\cite{huang2022real,hojjat2024limitnet}.

\begin{figure}[t]
\centering
\includegraphics[scale=0.5]{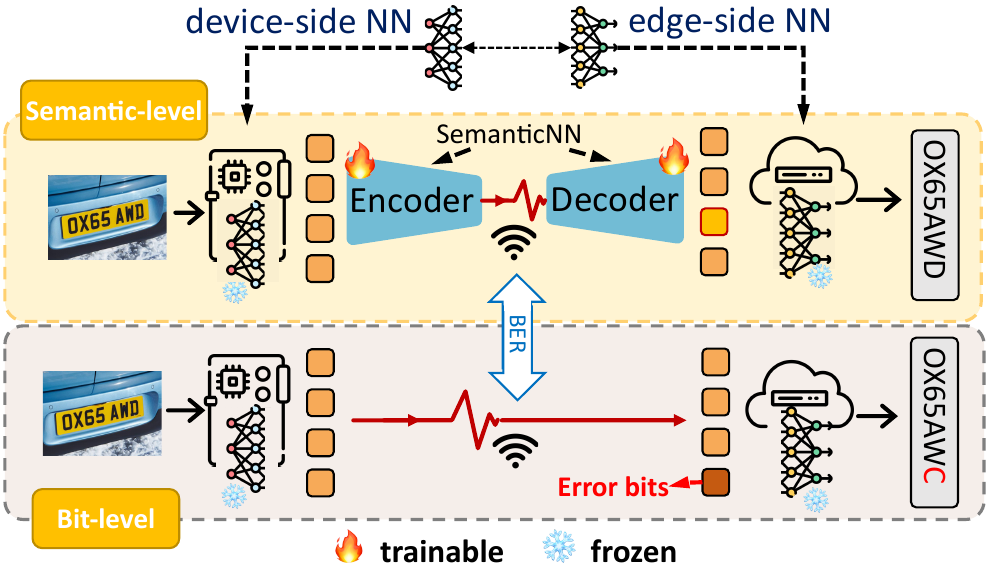}
\caption{SemanticNN (including encoder and decoder) accepts bit errors in wireless transmissions, pursuing semantic correctness with extremely feature compression in split computing. The original task NN will be frozen, and only the SemanticNN will be optimized.}
\label{semantic_inference}
\end{figure}


Nevertheless, executing AI algorithms on weak devices remains challenging due to severe constraints on computation, memory, and power~\cite{he2016deep, stm32datasheet}.
For example, ResNet-50 requires at least 100MB of memory, while a typical embedded device STM32, offers less than 1MB of local memory.
To bridge this huge gap, techniques such as TinyML, model compression, and pruning~\cite{kallimani2024tinyml,zhang2021mdldroidlite,lin2020hrank} have been widely adopted.
However, they may lead to non-negligible accuracy loss~\cite{frankle2020pruning, huang2022real} due to aggressive simplification of NN.
 For example, TinyML achieves over 50\% compression via low-precision quantization, but this comes at the cost of reduced model expressiveness and degraded task performance. 

Recent advances in split computing~\cite{kang2017neurosurgeon, laskaridis2020spinn,huang2021enabling, benazir2024speech, bin2024coacto} significantly mitigate accuracy degradation by offloading parts of NN computations from weak devices to edge servers.
Concretely, the original NN is split into the device-side NN and the edge-side NN, with the intermediate features extracted on the device offloaded to the edge for further inference.
Existing works primarily focus on optimizing model partitioning strategies and compressing device-side outputs to reduce both computational load and communication overhead.



However, these offloaded features are typically compressed before transmission to reduce bandwidth usage, making them highly vulnerable to bit errors introduced by wireless communication~\cite{nogales2018makersat,becke2012fairness}. 
Even a small number of flipped bits can lead to significant semantic drift in downstream predictions. For instance, causing a speech-to-text system to misinterpret “temperature” as “temperate”.
This sensitivity highlights the importance of transmission robustness in AI systems. 


Traditional bit-level error correction techniques, such as network coding\cite{wicker1999reed,mackay2005fountain} and retransmission protocols, are designed to ensure bit-level reliability for upper-layer applications.
However, they could be impractical for resource-constrained devices, particularly under extremely dynamic or unreliable wireless links \cite{yun2010towards}. 
Orthogonal to bit-level approaches~\cite{bragilevsky2020tensor,bajic2021latent,dhondea2021caltec,itahara2022communication,wang2022neuromessenger}, we propose a novel paradigm based on semantic-level correctness. 

In this paper, we introduce \emph{SemanticNN}, a semantic codec comprising an encoder and decoder, designed for error resilient \emph{Semantic} \emph{N}eural \emph{N}etwork offloading.
Unlike traditional bit-level approaches, SemanticNN treats bit errors as acceptable as long as the essential semantic content is preserved. 
As illustrated in Figure~\ref{semantic_inference}, our approach extends Shannon’s second level of communication~\cite{shannon1948mathematical} by defining semantic correctness in terms of end-to-end inference accuracy. 
The encoder compresses the output of the device-side NN, while the decoder reconstructs the received features in the presence of transmission errors. 
Thus, the primary objective of SemanticNN is to design an encoder-decoder framework capable of tolerating transmission errors. 
Two key challenges arise in designing SemanticNN.


The first stems from the combination of extreme feature compression and dynamic transmission errors.
Specifically, extreme feature compression increases sensitivity to transmission errors. 
Moreover, this sensitivity is further amplified by the varying severity of transmission errors (e.g., multipath interference in Wi-Fi), which hinder the accurate recovery of semantic information at the edge (Appendix~\ref{moti:coding}). 
To tackle this, we design a dedicated semantic codec consisting of a Soft Quantization (SQ)-based encoder and a BER-aware decoder.
The encoder introduces a learnable, non-uniform quantization mechanism that allows end-to-end optimization of both the quantization and the NN parameters. 
This ensures that even under severe compression, the encoder retains the most semantically meaningful representations.
The decoder integrates an adaptive attention module that dynamically responds to varying error rates, enabling robust reconstruction of corrupted features.
Furthermore, we propose Feature-Augmentation Learning, a two-stage training framework.
The first stage involves pre-training the codec using an autoencoder objective to denoise compressed features, serving as a universal initialization applicable to any downstream task at a given model split point. 
In the second stage, task-specific model fine-tuning refines the ability to capture high-level semantic features tailored to individual applications.


The second lies in the architectural asymmetry between the encoder and decoder.
Due to resource constraints on weak devices, the encoder deployed at the edge is significantly simpler in structure than the decoder on the server.
Our experiments in Appendix~\ref{sec:ays} reveal that the decoder often has over ten times the capacity of the encoder.
While deeper decoding models generally offer better decoding performance, the limited expressiveness of the lightweight encoder results in suboptimal feature extraction.
This architectural asymmetry significantly degrades the task performance.
To address this, we propose asymmetry compensation based on eXplainable AI (XAI) techniques for the encoder in SemanticNN.
By leveraging explainability maps generated through XAI techniques, the encoder identifies and prioritizes the most informative regions of the feature maps.
This lightweight strategy enables the encoder to enhance its semantic feature extraction capability without introducing additional computational or memory overhead.


Our main contributions can be summarized as follows:
\begin{itemize}
	\item We propose SemanticNN, a semantic codec for error resilient compressive device-edge collaboration, comprising SQ-based encoder and BER-aware decoder, trained through Feature-augmentation Learning. 
	\item We introduce XAI-based Asymmetry Compensation that enables the weak encoder in SemanticNN to refine feature map based on explainability analysis for extracting more valuable information without extra overhead.
	\item We conduct extensive experiments and the results show that SemanticNN drastically reduces the offloaded features by 56.82-344.83$\times$ while outperforming all baselines. Real-world case study further demonstrates its robustness to dynamic BERs.
\end{itemize}

 The rest is organized as follows. Section~\ref{related_work} reviews related work.
 Section~\ref{sec:overview} shows SemanticNN architecture and training process. 
 Section~\ref{tech2} details the encoder compensation mechanism.
 Section~\ref{evaluation} presents experimental results and real-world cases.
 Finally, Section~\ref{conclusion} concludes the paper.

\begin{figure*}[t]
	\centering
	\includegraphics[width=0.90\linewidth]{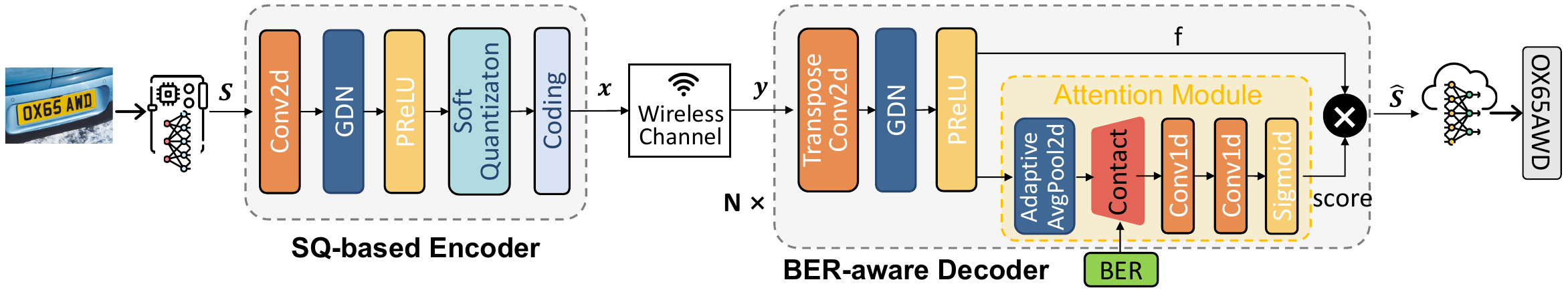}
	\caption{The detailed model structures of the SQ-based encoder and BER-aware decoder in SemanticNN.}
	\label{SemanticNN}
\end{figure*}

\section{Related Work}
\label{related_work}

\subsection{Efficient On-device Inference}
\textbf{Model Compression.}
To align the computational requirements of tasks with the weak devices, researchers have pursued model compression techniques~\cite{lin2020hrank,hinton2015distilling,sau2016deep}, which reduce model size and computational overhead while maintaining accuracy.
Alternatively, many lightweight architectures~\cite{sandler2018mobilenetv2} are designed directly to achieve efficiency without compression. 


However, when deploying models on weak devices, existing approaches often oversimplify the architecture, leading to significant accuracy degradation~\cite{huang2022real,zhuo2022empirical}. 

\textbf{Device-edge collaboration.}
%
Research on device-edge collaboration has explored optimization of computation allocation and communication overhead between the device and edge. 
For communication efficiency, techniques like semantic caching~\cite{xu2018deepcache}, input filtering, and semantic similarity indexing~\cite{yuan2022infi,kang2022tasti} have been proposed. 
For computation, early approaches~\cite{kang2017neurosurgeon,li2018jalad} introduced partial NN offloading to enable collaborative inference, paving the way for split learning. 
Subsequent work~\cite{laskaridis2020spinn,huang2021enabling}, building upon the early-exit mechanism~\cite{teerapittayanon2016branchynet}, adopted progressive inference to generate intermediate outputs. 
Further optimizations~\cite{yao2020deep,huang2022real} compressed intermediate data to enhance real-time performance. 

However, none of the aforementioned works take into consideration the fact that the potential occurrence of transmission errors can substantially affect task performance.


\subsection{Error-Resilient Communication}
\label{semantic_communication}

Shannon's theory of communication defines three hierarchical levels~\cite{shannon1948mathematical}. 
The first focuses on the \emph{physical} transmission of bits. 
The second emphasizes the correct conveyance of \emph{semantic} meaning. 
This paper leverages semantic-level communication for robust split AI inference.

\textbf{Bit-Level Reliable Communication}. 
A wide range of techniques have been developed to improve the reliability and efficiency of bit-level transmission, including sophisticated network coding and modulation schemes\cite{svensson2007introduction,sundararajan2008arq}. 
With the rise of AI, recent efforts have extended these communication principles for AI tasks. 
For example, tensor completion methods~\cite{bajic2021latent,dhondea2021caltec} aim to recover lost packets during transmission, while NeuroMessenger~\cite{wang2022neuromessenger} improves resilience under low SNR conditions through adaptive retransmission mechanisms. BottleNet++~\cite{shao2020bottlenet++} further integrates Joint Source Channel Coding (JSCC) into original AI models, enabling end-to-end optimization for error-resilient bit transmission.

However, in long-distance or interference-prone scenarios, achieving perfect bit-level fidelity often requires highly conservative coding strategies. 
These methods, while effective in minimizing transmission errors, can incur significant overhead due to the dynamic nature of wireless channels.

\textbf{Semantic-Level Reliable Communication.}  
In contrast to prior approaches prioritizing bit-level accuracy, semantic-level communication tolerates minor bit errors as long as the essential semantic meaning is preserved. 
Bit-level and semantic-level optimizations are orthogonal, i.e., combining them can yield complementary benefits~\cite{shao2021learning,lee2019deep}.
%
Recent advances leverage deep learning (DL) to extract semantic information directly from data, forming the basis of semantic communication\cite{bourtsoulatze2019deep,xie2021deep}. 
Most DL-based frameworks adopt the JSCC paradigm, which jointly optimizes source encoding, channel characteristics, and task-specific decoding. 
This design enables efficient transmission of multimodal data under poor channel conditions \cite{xie2021deep,xu2021wireless,weng2023deep,han2022semantic,dai2022nonlinear}.

SemanticNN aligns with this trend, but introduces that instead of transmitting raw data, semantic communication can be applied to intermediate feature maps in split computing. 
This approach enhances error resilience and supports extreme compression for deploying AI on weak devices.

\section{Building the Model}
\label{sec:overview}
%
In this paper, we focus on resource-constrained devices and fix the split after the first convolutional layer.
Given the large number of possible split positions and the fact that earlier splits cause greater performance degradation (Appendix A.2), this configuration represents the most challenging scenario for compressive offloading under transmission errors.


\subsection{Model Structure}
The model architecture of SemanticNN forms the foundation for achieving efficient and error-resilient semantic offloading. 
As illustrated in Figure~\ref{SemanticNN}, it consists of an SQ-based encoder and a BER-aware decoder.

\subsubsection{SQ-based Encoder}
The \emph{SQ-based encoder} is responsible for compressing feature representations under extreme resource constraints.
It consists of one convolutional layer, followed by Generalized Divisive Normalization (GDN), Parametric Rectified Linear Unit (PReLU), and a learnable soft quantization layer coupled with fixed-length bit coding.

\emph{NN Operations.}
GDN is a normalization technique, and it has been widely adopted in perceptual compression tasks due to its effectiveness in capturing statistical dependencies among neural activations. 
PReLU further improves the expressiveness of the encoder by mitigating the \emph{dying ReLU} problem, allowing a small gradient for negative inputs to maintain neuron activity during training.

\emph{Feature Compression Mechanism.}
The soft quantization and bit coding perform further compression on extracted features to reduce transmission overhead. 
Motivated by findings in Appendix~\ref{moti:coding}, we adopt fixed-length coding to enhance stability and avoid error propagation caused by variable-length codes like Huffman coding.

\begin{figure}[t]
\centering
\includegraphics[width=\linewidth]{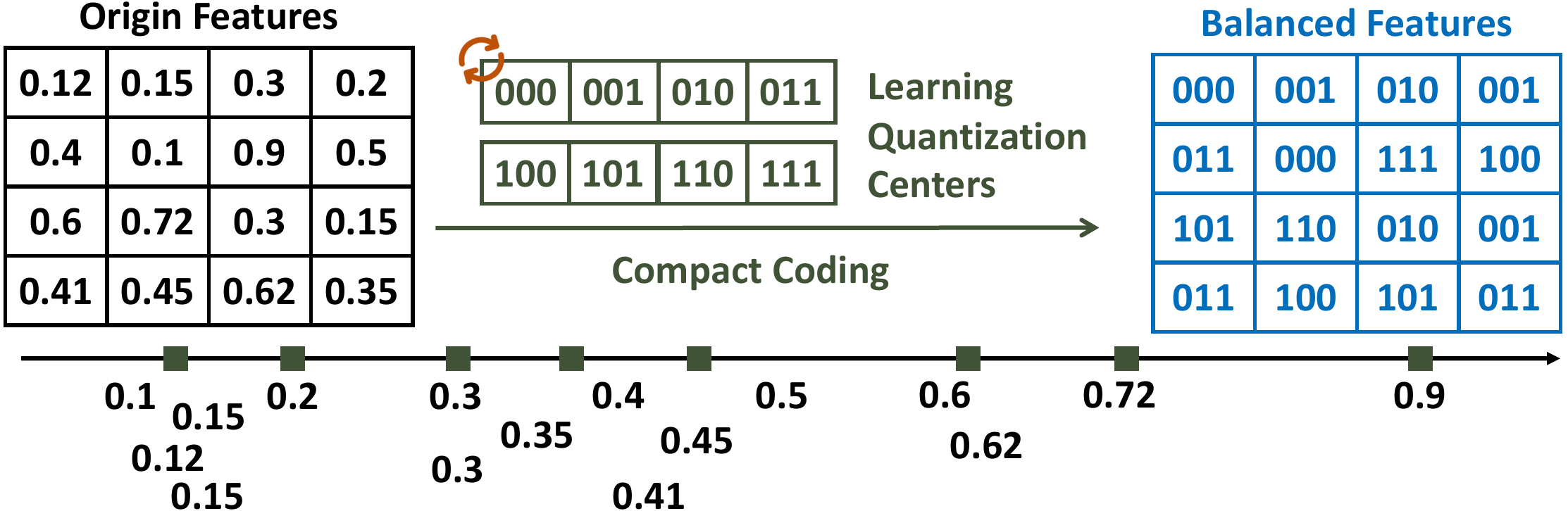}
\caption{Balanced quantization with eight learning quantization centers and 3-bit compact coding.}
\label{quantization}
\end{figure}

As illustrated in Figure~\ref{quantization} (left), traditional hard (non-uniform) quantization is employed to partition the feature space into intervals, each represented by a centroid.
However, it introduces two major limitations: (1) suboptimal centroid placement when dealing with skewed or outlier-prone distributions, and (2) it is non-differentiable, which blocks gradient propagation during training.
To address these issues, we introduce a learnable soft quantization mechanism that enables end-to-end optimization of both the quantization parameters and the overall network. 
As shown in Figure~\ref{quantization}, the encoder learns eight quantization centers, each representable using three bits, through backpropagation.
This quantization process is treated as a differentiable neural layer, allowing joint optimization of model parameters and data distribution statistics.

%

\subsubsection{BER-aware Decoder}
The \emph{decoder} comprises $N$ blocks of similar operations to the encoder supplemented by an attention module each.
Corresponding to the encoding operation, we utilize transpose convolution and inverse GDN  to reconstruct the intermediate feature maps during the decoding process.
The \emph{attention module} in the decoder calculates the attention score to quantify the level of attention paid to intermediate features under the prevailing noise conditions.
It accepts input from the intermediate features and the current communication state (i.e., BER), learning the correlation between the transmission error and the reconstructed feature. 
With this attention module, SemanticNN to dynamically adapt to unpredictable changing communication link environments.
Through iterative repetitions of this process, the intended offloaded feature map will be restored at the receiver size.
Finally, the more complex trained decoder will be deployed to edge servers as the receiver.

\subsection{Feature-augmentation Learning}
The designed model needs to be trained to compress offloaded activations efficiently while preserving semantic features crucial for task performance.
We propose progressive Feature-augmentation Learning. 
It first trains the denoising autoencoder to achieve initial noise reduction while reducing the possibility of overfitting for downstream tasks. 
Subsequently, Task-oriented semantic-level training is performed on the downstream task to extract semantic information while adapting to transmission errors.
Figure~\ref{overview} shows the details of the training process.

\subsubsection{Denoising Autoencoder Training}
\label{stage1}
\begin{figure}[t]
\centering
\includegraphics[scale=0.36]{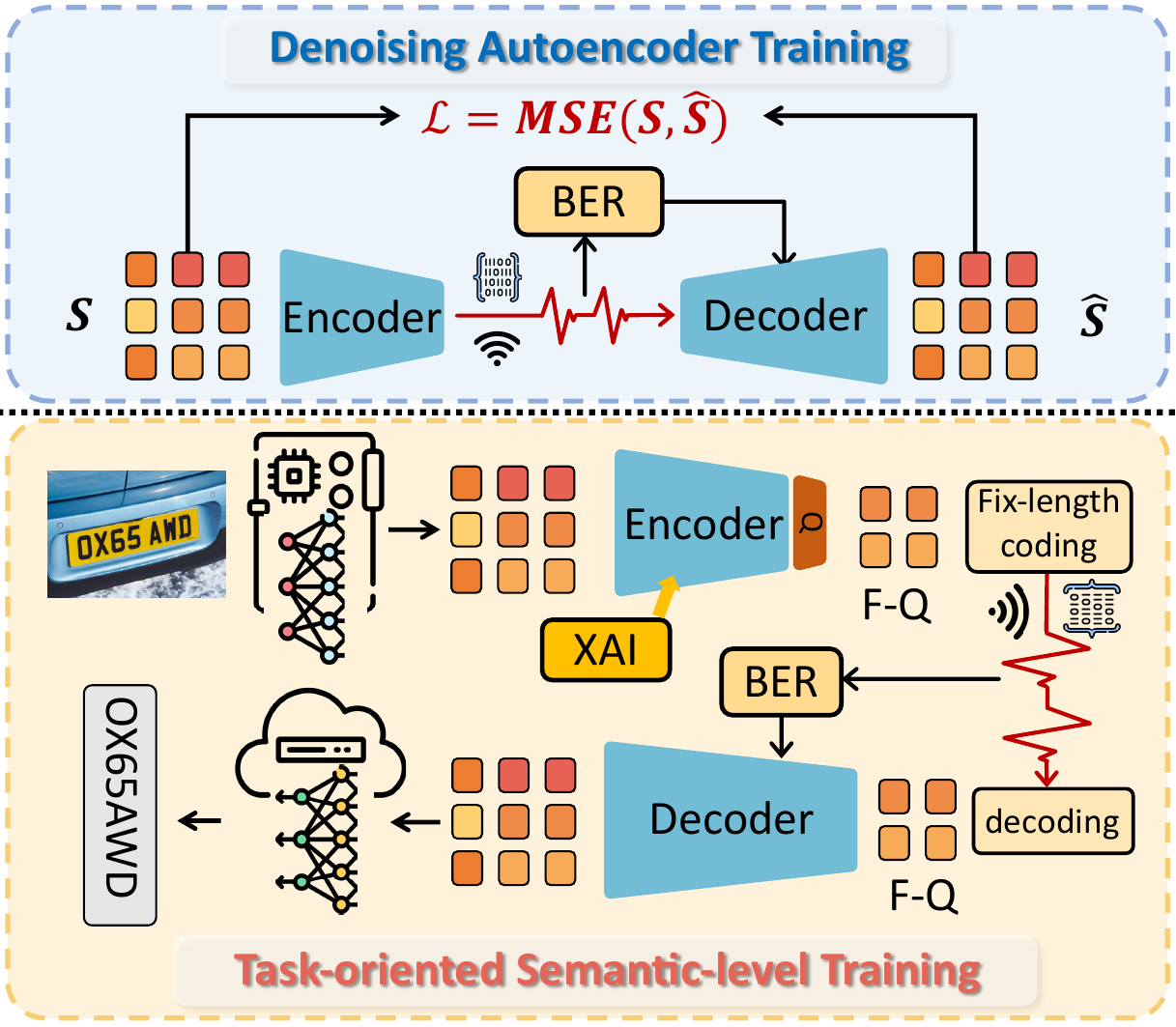}
\caption{Feature-augmentation Learning first performs Denoising Autoencoder Training to initially fight against transmission errors. Subsequently, Task-Oriented Semantic-level Training is performed to extract and transmit semantic information on specific downstream tasks for better performance.}
\label{overview}
\end{figure}

This phase aims to align transmitted and received features at the bit level, enhancing resilience to counter interference caused by transmission errors.
As illustrated on the top of Figure~\ref{overview}, we introduce Mean Squared Error (MSE) to minimize the difference as shown in Equation~\ref{mse}:
\begin{equation}
\label{mse}
\mathcal{L}_{\text{bit}} = \text{MSE}\left(\mathbf{s}, \hat{\mathbf{s}}\right).
\end{equation}

As training progresses and $\mathcal{L}_{\text{bit}}$ diminishes, the encoder and decoder gradually converge toward a consistent representation of the input features, achieving bit-level alignment. This pre-training stage enables SemanticNN to develop an initial resilience to transmission errors, endowing it with error-mitigation capabilities even on unseen tasks. Furthermore, this process significantly enhances the learning efficiency of downstream task adaptation while reducing the risk of overfitting during semantic feature extraction. Notably, this phase is entirely task-agnostic and depends only on the size of the intermediate feature maps at the offloading point. Therefore, encoders and decoders trained under this phase can be directly reused across different tasks, as long as the feature dimensions match.


\subsubsection{Task-Oriented Semantic-level Training}
\label{stage2}
At this phase, we draw on the idea of semantic communication to 
transmit the most task-relevant information with a degree of tolerance for bit errors for better task performance.
SemanticNN and the origin NN model are both considered for the Task-oriented Semantic-level Training, which achieves the goals of error-resilient compressed semantic inference. 
The origin NN model is solely engaged during the training process to produce intermediate features and does not participate in the subsequent gradient updates.
The performance of tasks is safeguarded by aligning the offloaded inference results at the edge with the ground truth.
It enables the transmission of the most task-relevant data.
To further improve the efficiency of fixed-length coding, we incorporate a distribution regularization term into the loss function. 
Given the task-dependent nature of feature distributions, we define the semantic-level training objective as:
\begin{equation}
\label{low_entropy_loss}
\mathcal{L}_{\text{sem}} = 
\alpha \cdot \mathcal{L}_{\text{div}}\left(\mathbf{p}, \frac{1}{n}\mathbf{1}\right) + 
\beta \cdot \mathcal{L}_{\text{cls}}(y, \hat{y}) + 
\gamma \cdot \mathcal{L}_{\text{XAI}},
\end{equation}
where $n$ is the number of quantization centers and $\mathbf{p} \in \mathbb{R}^n$ denotes the normalized frequency distribution of their occurrences. 
The distribution regularization loss $\mathcal{L}_{\text{div}}$ encourages frequency distributions to approach a uniform distribution.
The task-specific loss $\mathcal{L}_{\text{cls}}(y, \hat{y})$, such as cross-entropy (CE) for classification or MSE for regression, measures the discrepancy between the ground truth $y$ and the predicted output $\hat{y}$. 
The $\mathcal{L}_{\text{XAI}}$ term, introduced in Section~\ref{sec:loss}, promotes semantic interpretability by encouraging sparse and faithful feature attributions. 
Together, the multi-objective formulation balances model performance, feature utilization, and explainability. 
Hyperparameters are justified in Appendix~\ref{param_study}. 

Adapting to transmission errors in denoising autoencoder training enhances the efficiency of semantic extraction and transmission in this stage.
Bypassing the first stage and directly imposing this constraint on SemanticNN may lose the fight against transmission errors, potentially introducing instability.
Appendix~\ref{two_stage} shows the results of ablation studies.
To further improve the ability of the asymmetric encoder in SemanticNN, we impose an additional loss constraint $\mathcal{L}_{\text{XAI}}$ in Equation~\ref{low_entropy_loss}, which will be introduced in Section~\ref{tech2}.

\section{XAI-based Asymmetry Compensation}
\label{tech2}
\begin{figure}[t]
\centering
\footnotesize
\includegraphics[scale=0.29]{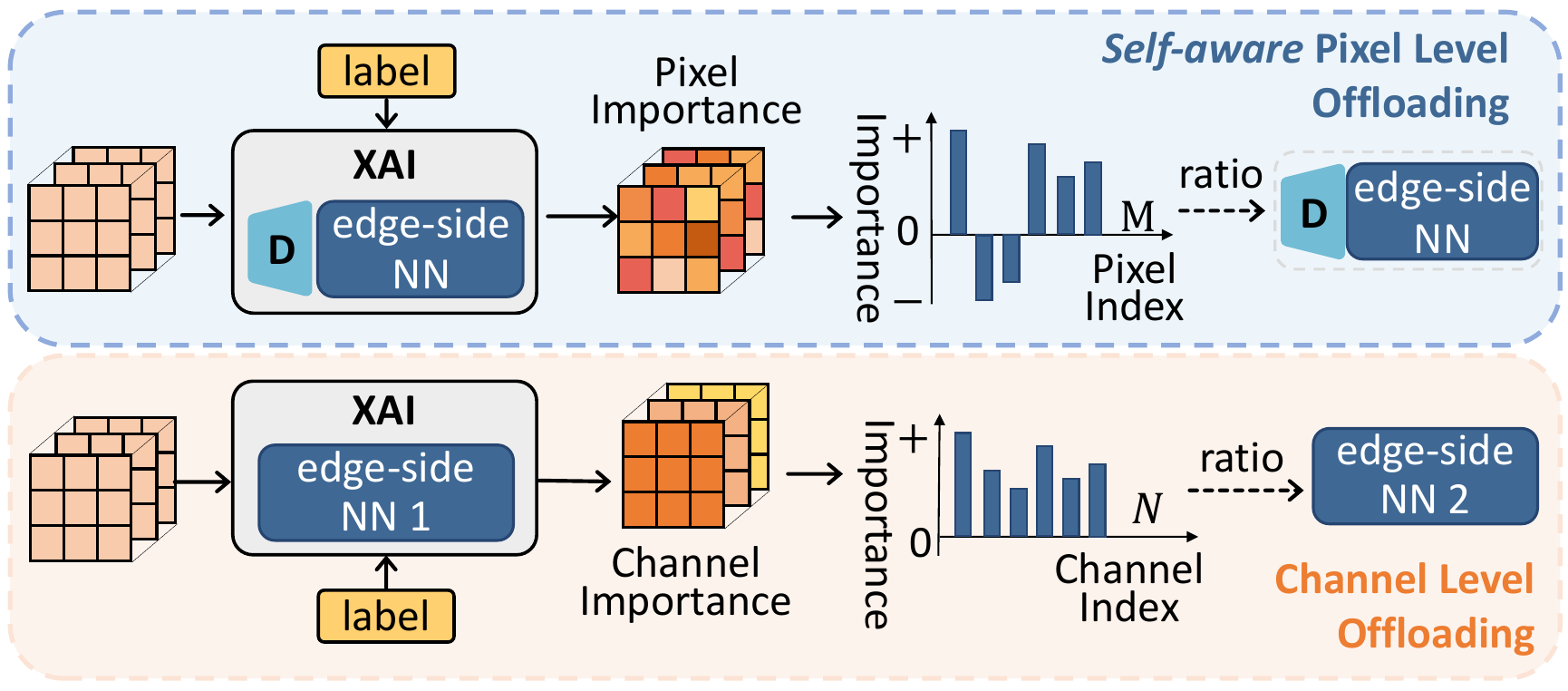}
\caption{Self-aware pixel-level feature importance analysis using XAI and partial feature offloading.}
\label{xai}
\end{figure}

\emph{XAI-based Asymmetry Compensation} is primarily employed to optimize the performance of the asymmetric device-side encoder.
We propose to enhance the encoder by extracting important semantic information and further compress the transmission data by removing non-important features with the help of external XAI tools, shown at the top of Figure~\ref{xai}.


\label{sec:loss}
Channel-level analysis~\cite{huang2022real} utilizes XAI tools to differentiate between important and unimportant features at the channel level in task offloading scenarios.
In contrast to conducting feature map importance analysis at the channel level (the bottom of Figure~\ref{xai}), SemanticNN proposed \emph{pixel-level} feature map importance analysis, motivated by Appendix~\ref{sec:mot_pix}.


The top of Figure~\ref{xai} illustrates the detailed utilization of XAI tools in the training of the encoder and the decoder within SemanticNN. 
In contrast to channel-level analysis, SemanticNN involves pixel-level analysis, aiming to transmit crucial pixels that hold significance across multiple channels during the training process.
SemanticNN harnesses XAI in two facets: firstly, to transmit fewer data incorporating a maximal number of important features, and secondly, to optimize the significance of these important features.
This optimization of importance is shown as Equation~\ref{xai_loss}, which is also added on Equation~\ref{low_entropy_loss}:
\begin{equation}
\label{xai_loss}
\mathcal{L}_{\text{XAI}} = 
\lambda \cdot \frac{1}{\sum_{i} \max(\phi_i, 0)} 
+ (1 - \lambda) \cdot \frac{M}{|\{i : \phi_i > 0\}|} 
- r,
\end{equation}
where $M$ is the total number of offloaded features, 
$\sum_{i} \max(\phi_i, 0)$ is the sum of positive attribution scores, 
$|\{i : \phi_i > 0\}|$ is the number of features with positive attributions, 
and $r$ is a scaling factor that ensures the reward term is on a comparable scale. 

%
%

\begin{algorithm}[t]
\small
\renewcommand{\algorithmicrequire}{\textbf{Input:}}
\renewcommand{\algorithmicensure}{\textbf{Output:}}
\caption{Pixel-Level Feature Slicing}
\label{slicing}
\begin{algorithmic}[1]
\Require Intermediate feature map $\mathbf{F} \in \mathbb{R}^{C \times H \times W}$, spatial ratio $\alpha \in (0,1]$.
\Ensure Cropped feature map $\mathbf{F}_\alpha$ of size proportional to $\alpha$.

\State $H \gets \text{height}(\mathbf{F})$, $W \gets \text{width}(\mathbf{F})$
\State $S_{h}, S_{w} \gets \left\lfloor \sqrt{\alpha}, H \right\rfloor, \left\lfloor \sqrt{\alpha}, W \right\rfloor$
\State $p_{h}, p_{w} \gets \left\lfloor \frac{H - S_{h}}{2} \right\rfloor, \left\lfloor \frac{W - S_{w}}{2} \right\rfloor$
\State $\mathbf{F}_\alpha \gets \mathbf{F}[:, p_h : p_h + S_h, p_w : p_w + S_w]$
\State \textbf{return} $\mathbf{F}_\alpha$
\end{algorithmic}
\end{algorithm}

SemanticNN aims to offload all balanced quantified features.
Additionally, it also enables partial offloading (i.e., offloading features of a specified ratio to the edges) to adapt to constraint communication resources.
We utilize pixel-level feature slicing as depicted in Algorithm~\ref{slicing}.
For all channels, we cut off the corresponding percentage of pixels.
For slice ratio, it is best not to exceed 50\%, and as Figure~\ref{channel_level} in the Appendix reveals, around 50\% features are important.
Correspondingly, at the receiver end, SemanticNN employs the image optimization method known as \emph{ReflectionPad2d}~\cite{lin2023comparative} for recovering from the untransmitted pixels, which is more effective compared to traditional zero padding.

Besides, the most important component is the guidance model in XAI, which is used to characterize the importance of features.
Indeed, different models assess the significance of features within the same input in varying ways.
Therefore, instead of relying on another more potent model to ascertain input importance (i.e., edge-side NN 1 in Figure~\ref{xai}), we leverage the edge-side NN itself.
With Equation~\ref{xai_loss} added in $L_{sem}$, the entire NNs (i.e., original task models and SemanticNN) engaged in the end-to-end training process possess their interpretations of the input samples. 
\begin{figure*}[t]
\centering
	\begin{minipage}[t]{0.48\linewidth}
        \centering
        \includegraphics[width=\textwidth]{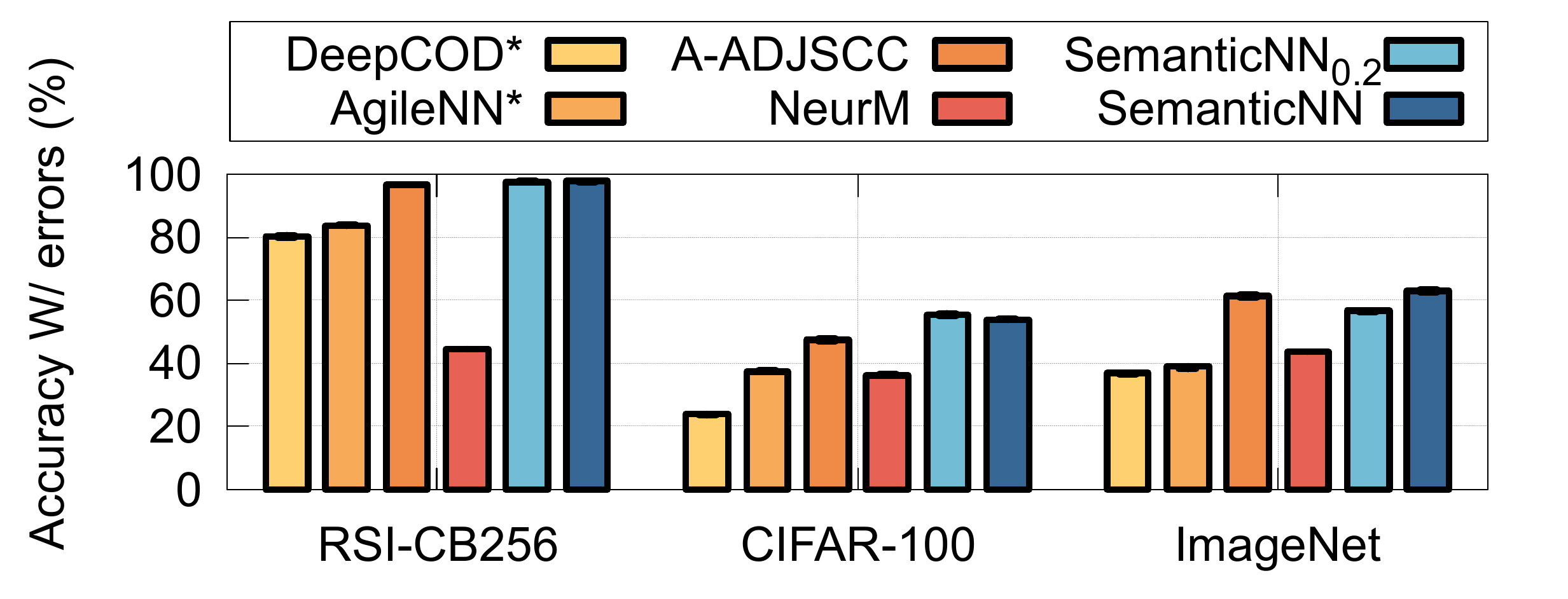}
        \centerline{(a) Accuracy of ResNet-50}
    \end{minipage}%
    \begin{minipage}[t]{0.48\linewidth}
        \centering
        \includegraphics[width=\textwidth]{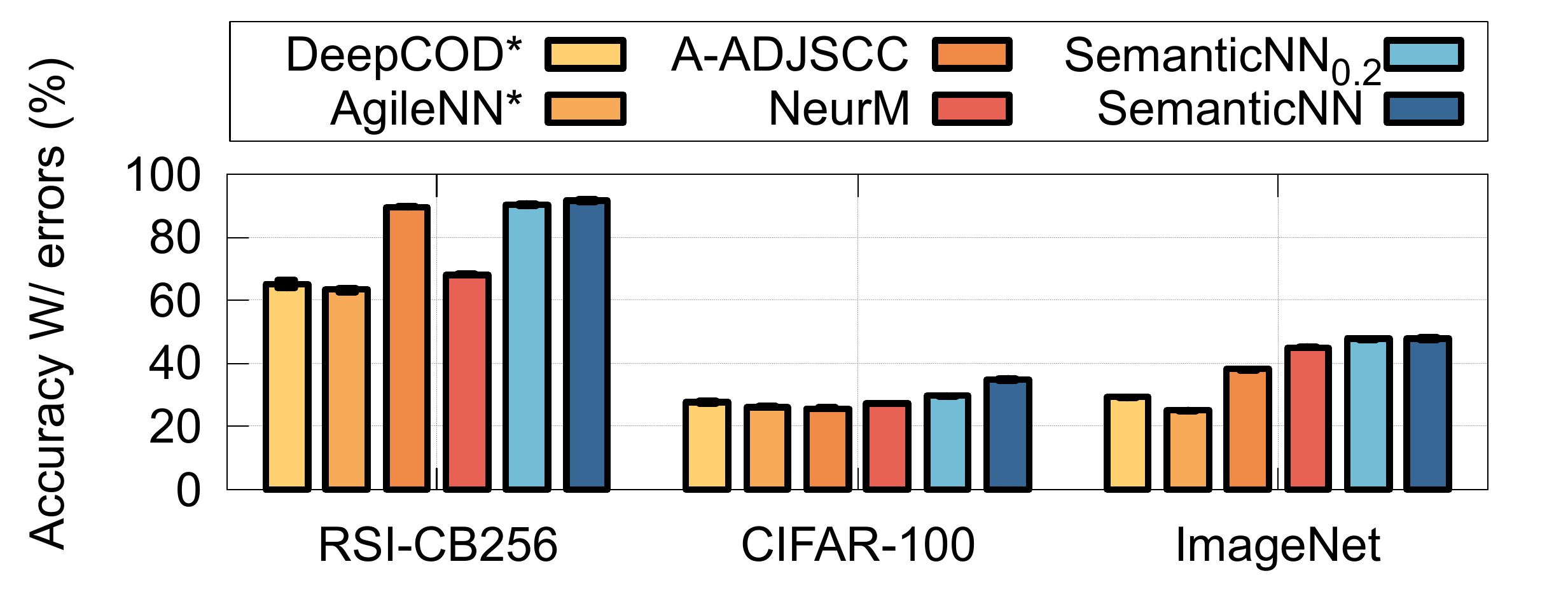}
        \centerline{(b) Accuracy of MobileNetV2}
    \end{minipage}
    \begin{minipage}[t]{0.48\linewidth}
        \centering
        \includegraphics[width=\textwidth]{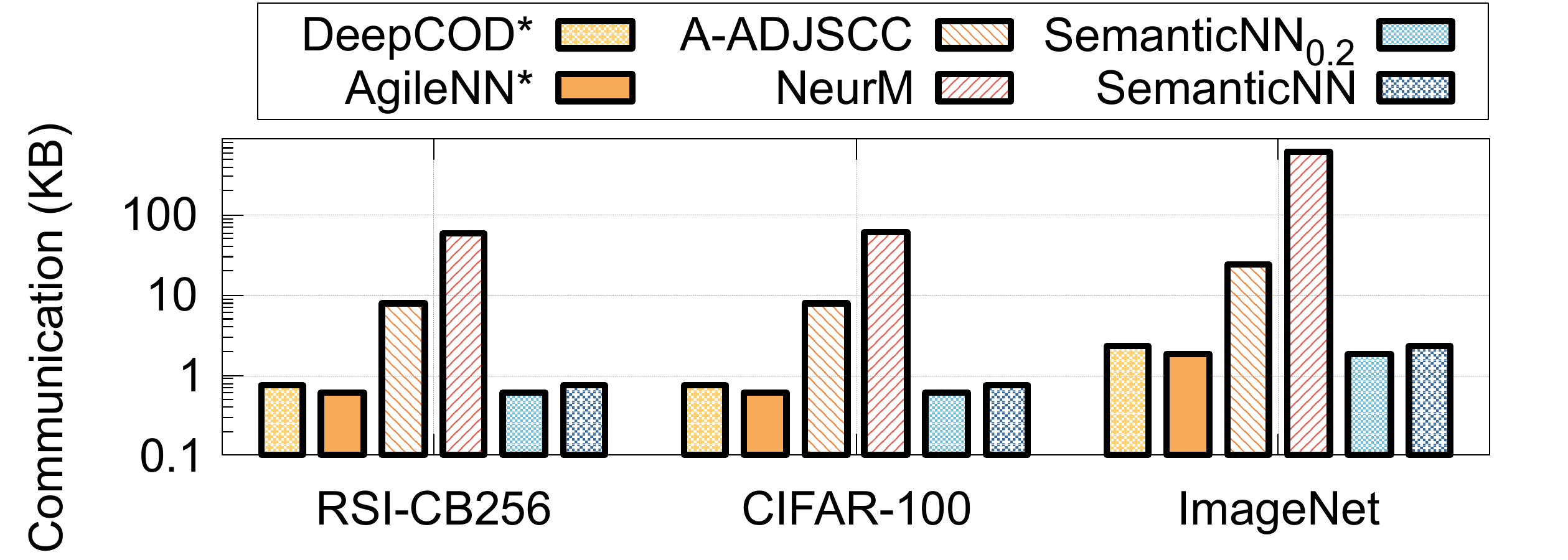}
        \centerline{(c) Offloaded features of ResNet-50}
    \end{minipage}%
    \begin{minipage}[t]{0.48\linewidth}
        \centering
        \includegraphics[width=\textwidth]{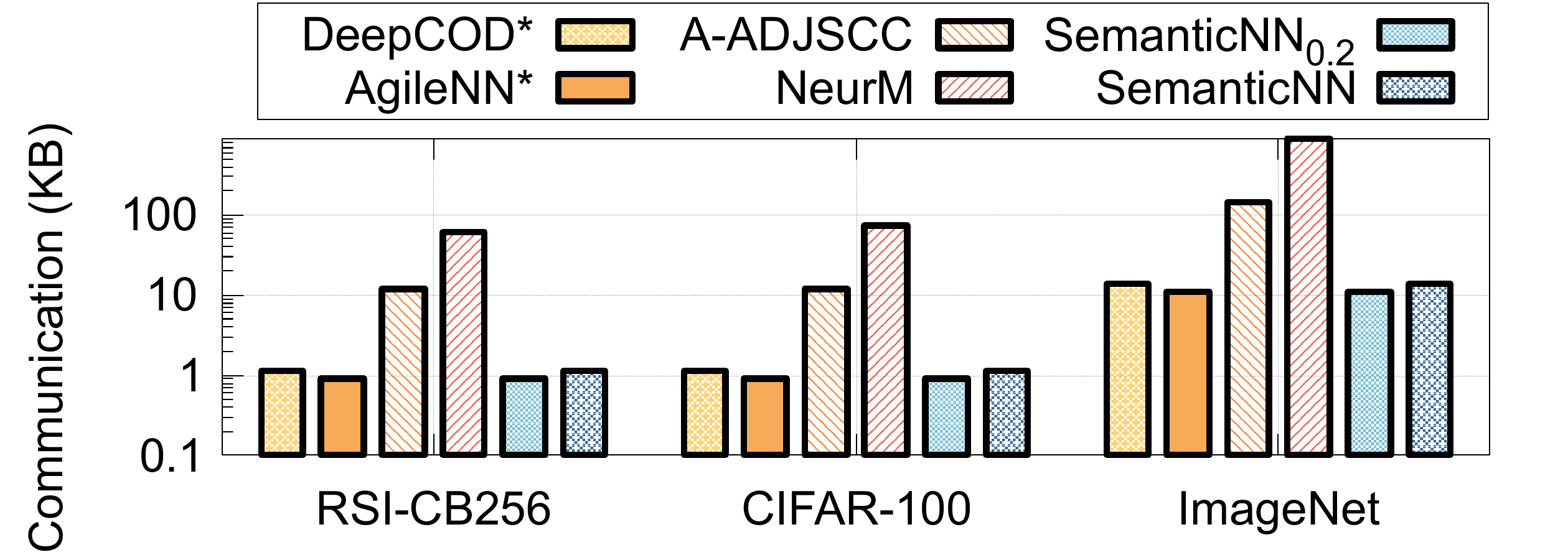}
        \centerline{(d) Offloaded features of MobileNetV2}
    \end{minipage}
\caption{Overall performance of image classification. SemanticNN$_{0.2}$ and SemanticNN represent two versions, where SemanticNN$_{0.2}$ offloads 20\% less amount of data than that of SemanticNN. SemanticNN achieves the best top-1 accuracy and less communication overhead with transmission errors across two pre-trained models and three datasets. We repeated 6 times, and the corresponding error bar indicates the stable task performance.}
\label{overall performance}
\end{figure*}


\section{Evaluation}
\label{evaluation}

\subsection{Datasets and setup}
\label{setup}
We implement SemanticNN with \emph{PyTorch}.
All models are trained on NVIDIA RTX-3090Ti GPUs and tested on STM32.
We manually inject bit-flips to the offloaded features for evaluation under various BERs, acknowledging the complexity of controlling BER in real-world scenarios.
According to real-world cases (Appendix~\ref{sec:case}), we set the BER range to be 0.01-5\% randomly. 
Due to space limitations, we have added the details of our experimental setup, datasets, metrics, and baselines in Appendix~\ref{sec:evalution_d}.

\begin{table}[t]
    \centering
    \small
    \begin{tabular}{c|ccc}
        \toprule
        \textbf{w/ E Performance}  & \textbf{Image} & \textbf{Feature} & \textbf{SemanticNN} \\
        \midrule
        \midrule
        RSI-CB256 (\%) & 54.76 & 59.28 & \textbf{97.92}\\  
        CIFAR-100 (\%)& 45.43& 29.32 & \textbf{53.96} \\   
        \midrule
        Data (KB)  & 48 & 64 &  \textbf{0.75}\\	
        \midrule	
        \midrule 
        ImageNet-200 (\%)  & 44.24 & 32.84 & \textbf{62.93} \\       
        \midrule
        Data (KB)& 588 & 748 & \textbf{2.3} \\		
        \midrule
        \midrule
        COCO (mAP) & 0.28 & 0.48 & \textbf{0.55}\\  
        VOC2007 (mAP) & 0.44 & \textbf{0.81} & 0.80\\    
        VHR (mAP) & 0.01  & 0.68 & \textbf{0.81} \\       
        \midrule
        Data (MB) & 4.69 & 1.17 & \textbf{0.037}\\	
        \bottomrule
    \end{tabular}
    \caption{SemanticNN achieves outperforming classification performance with transmission errors while the amount of offloading data is only 0.39-1.56\% of the original image and 0.31-1.17\% of features in size. It also achieves satisfactory object detection performance while offloading only 0.79\% of the original image.}
    \label{overall performance non-split}
\end{table}


\subsection{Performance for image classification} 
We compared the SemanticNN based on two base models with all baselines shown in Figure~\ref{overall performance} for top-1 accuracy with transmission errors and communication overhead (i.e., features per image to be offloaded). 

First, A-ADJSCC achieves slightly lower accuracy compared with two versions of SemanticNN, and outperforms the first two compressed baselines significantly.
Since A-ADJSCC needs more than 10$\times$ the amount for offloading due to the absence of quantization and coding strategies, compared with SemanticNN and the first two compressed baselines.
NeuroMessenger has a retransmission mechanism without quantization and coding strategies, resulting in the most communication overhead.

In split computing scenarios in IoT applications, the amount of offloaded data could be an important performance metric since the communication resource could be very constrained.
In split computing, the amount of offloaded data is a vital metric due to potentially limited communication resources.
The first two compressed baselines, i.e., DeepCOD* and AgileNN*, offload a similar amount of data compared with two versions of SemanticNN, which is only about 0.39\% to 2.34\% of the original image size. 
SemanticNN significantly outperforms the two baselines by 7.11\% to 30.88\% across various pre-trained models and datasets, showing the error resilience inference ability of SemanticNN.
AgileNN is the best existing approach in terms of this metric, by proposing an XAI-based feature importance manipulation scheme.
To compare the performance of AgileNN and SemanticNN fairly, SemanticNN$_{0.2}$ in Figure~\ref{overall performance} is a version of SemanticNN which ensures the same amount of offloaded data compared with that of AgileNN.
From the figure, we can see that although the accuracy of SemanticNN$_{0.2}$ is slightly lower than the accuracy of SemanticNN, SemanticNN$_{0.2}$ still outperforms AgileNN* significantly with the same amount of offloaded data.
Despite the streamlined inference efficiency of MobileNetV2, it falls short of the accuracy of ResNet-50, although it sustains a competitive compression rate.

The other scenario is offloading without any processing, as shown at the top of Table~\ref{overall performance non-split}.
SemanticNN outperforms these two baselines facing transmission errors.
From the results, we can see that SemanticNN performs at least comparable with them, offloading only 0.39-1.56\% of the original image and 0.31-1.17\% of features in size.
Moreover, considering realistic BER variations, real-world case studies further validate its effectiveness in Appendix~\ref{sec:case}.

\begin{figure}[t]
    \begin{minipage}[t]{0.5\linewidth}
        \centering
        \includegraphics[width=\textwidth]{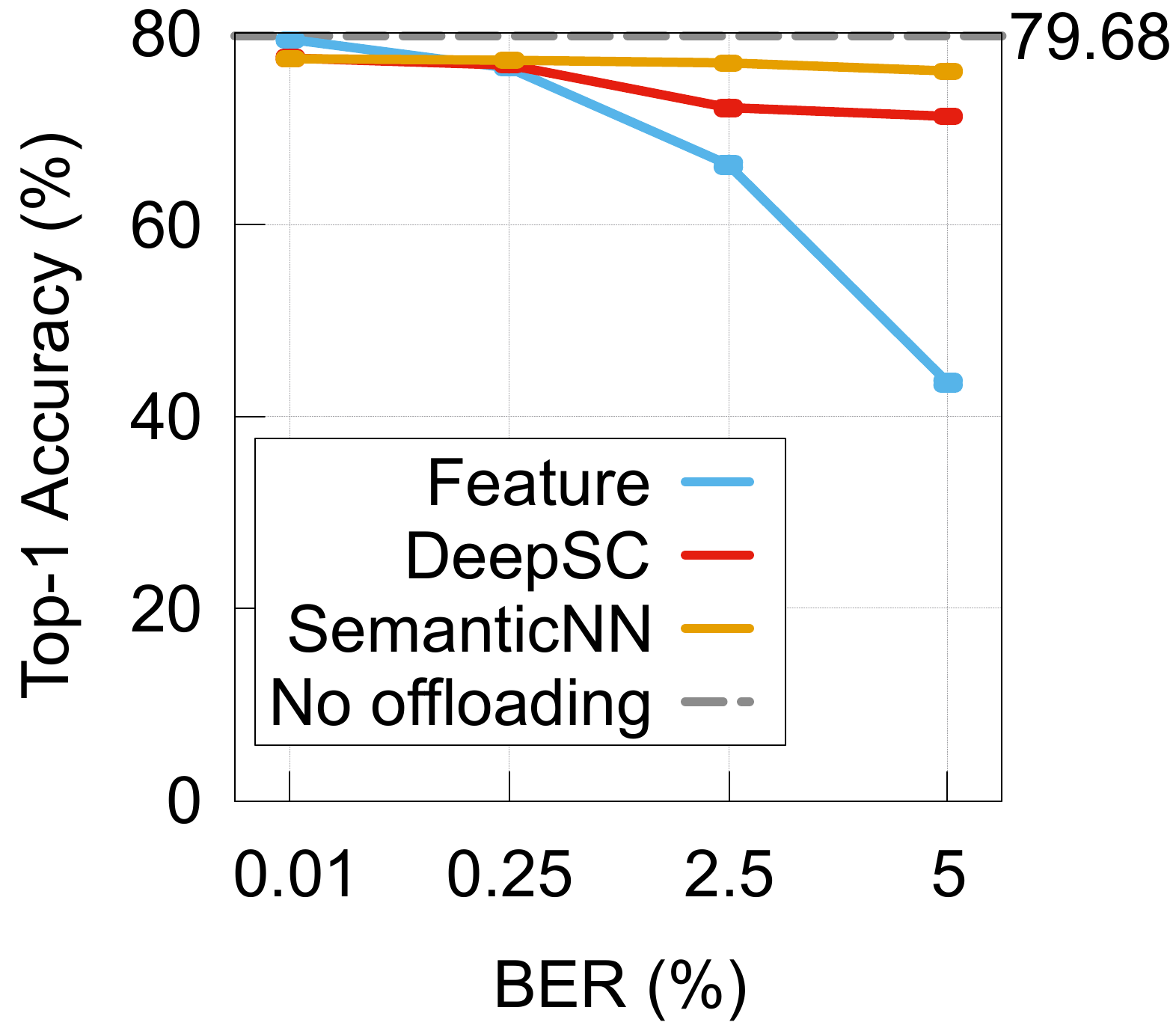}
        \centerline{(a) Under various BERs}
    \end{minipage}%
    \begin{minipage}[t]{0.5\linewidth}
        \centering
        \includegraphics[width=\textwidth]{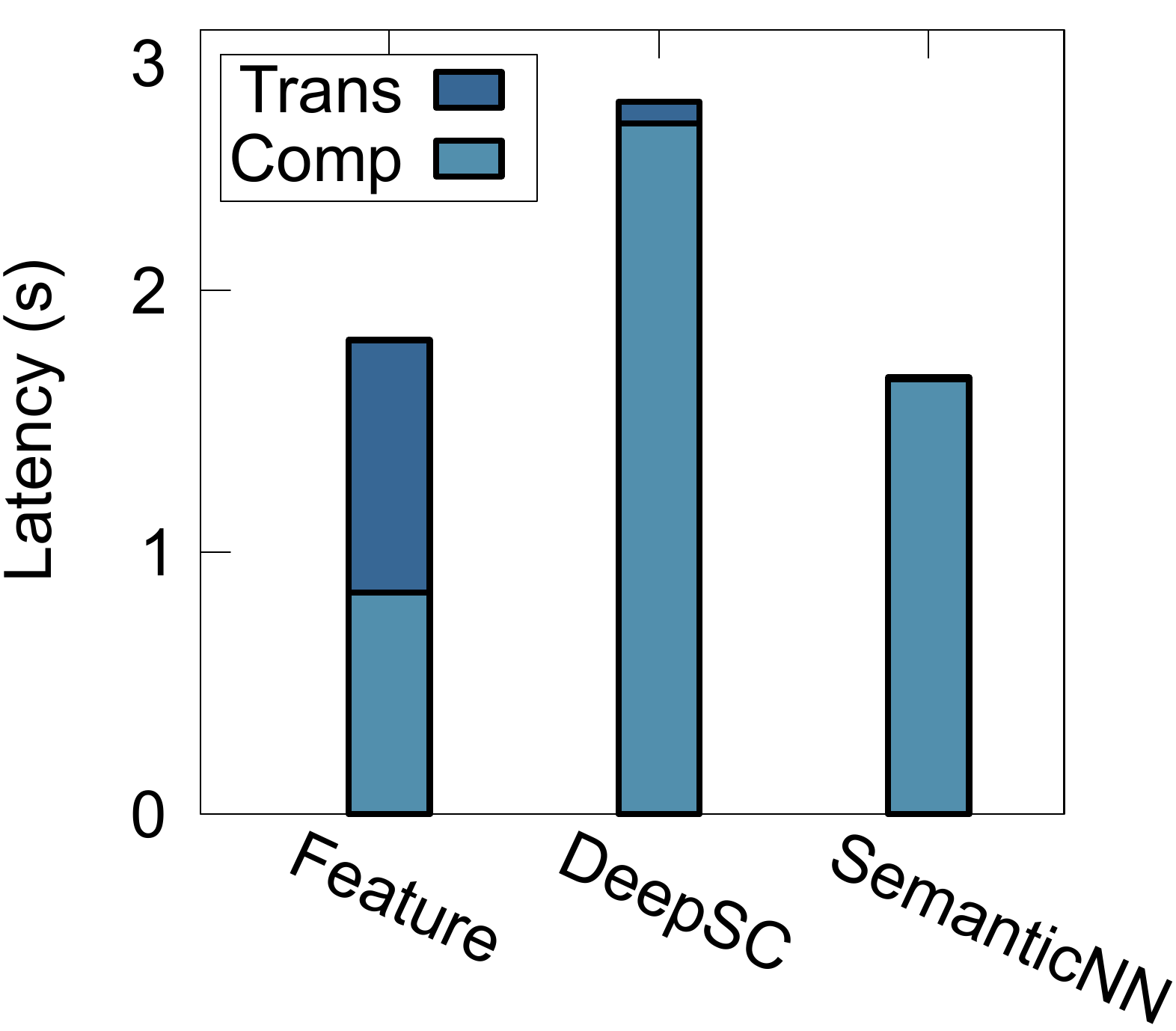}
        \centerline{(b) Latency}
    \end{minipage}

    \caption{CNN-based SemanticNN achieves better performance compared to transformer-based structures.}
    \label{model_structure_acc}
\end{figure}

\begin{table}[t]
    \centering
    \small
    \begin{tabular}{cccc}
        \toprule
        \multicolumn{2}{c}{\textbf{Params (KB)}}& \multicolumn{2}{c}{\textbf{Transmitted data (KB)}}\\
        \cmidrule(r){1-2} \cmidrule(r){3-4}
         DeepSC & SemanticNN & DeepSC & SemanticNN\\
         9165 & 1875 & 98 & 24.5\\		
        \bottomrule
    \end{tabular}
    \caption{SemanticNN utilizes a more lightweight CNN structure and achieves efficient data transmission compared to Transformer-based DeepSC.}
    \label{structures}
\end{table}

\begin{figure}[t]
\centering
    \includegraphics[scale=0.15]{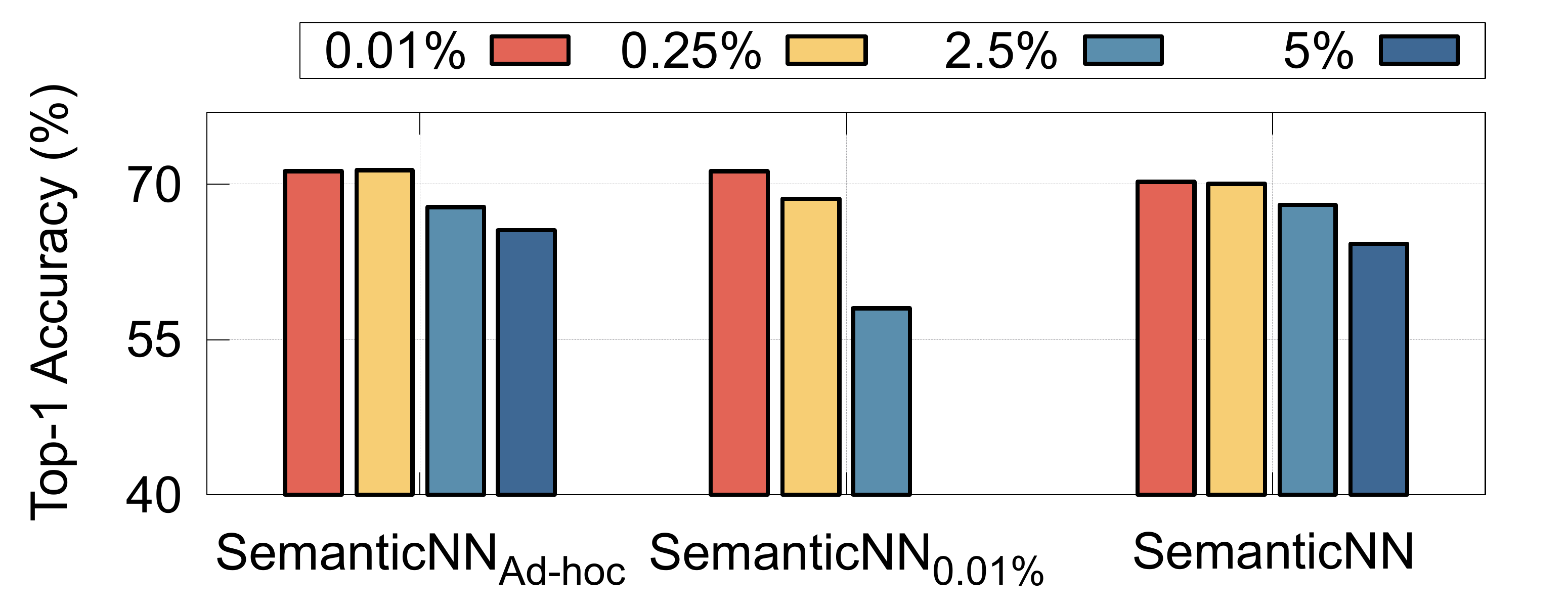}
    \caption{SemanticNN trained with attention under dynamic changes in BERs, achieves comparable accuracy as training SemanticNN at a specific BER.}
    \label{all_in_one}
\end{figure}

\subsection{Performance for object detection} 
We further evaluated SemanticNN using YOLOv3 on three varied datasets.
Since the baselines used in the image classification scenario did not include open-source implementation for object detection tasks, 
we only compare SemanticNN with basic methods, i.e., offloading the original image or feature map directly without any compression.
The bottom of Table~\ref{overall performance non-split} shows that SemanticNN also achieves satisfactory results with transmission errors.
Specifically, with only offloading data, which is only 0.79\% of the image size or 3.16\% of the feature size, Semantic achieves much higher accuracy with various transmission errors. 
Thus, SemanticNN demonstrates its versatility by being amenable to a broad spectrum of AI applications.

\subsection{Impact of Model Structures}

\noindent\textbf{Transformers vs. CNN.}
A large amount of research focused on semantic communication relies on \emph{Transformers} to enhance the extraction of semantic information from transmitted data.
Thus, we conducted a performance comparison with DeepSC, which employs a Transformer structure with two encoders and two decoders, across various BERs.
Continuing with the split computing of ResNet-50 on ImageNet-200, Figure~\ref{model_structure_acc} (a) illustrates the performance under various BER conditions: 0.01\%, 0.25\%, 2.5\%, and 5\%. 
With increasing BERs, SemanticNN achieves comparable performance with DeepSC.
However, for MobileNetV2 on CIFAR-100, Figure~\ref{model_structure_acc} (b) shows that
SemanticNN significantly minimizes transmission latency. 
Table~\ref{structures} shows the parameters of DeepSC and SemanticNN. 
Moreover, as for the overhead of weak edge devices, SemanticNN can achieve efficient data transmission with a more lightweight structure and less memory usage measured in Appendix~\ref{sec:overhead}.


\noindent\textbf{Attention in decoder.} 
SemanticNN$_{Ad-hoc}$ in Figure~\ref{all_in_one} represents training and testing at a specific BER. 
SemanticNN$_{0.01\%}$ represents testing at other BERs with a model trained at 0.01\% BER.
SemanticNN is trained at dynamically varying BERs and tested at a specific BER.
Results show that SemanticNN with attention modules can adapt to varying BERs and achieve comparable accuracy as training at a specific BER.

\section{Conclusions}
\label{conclusion}

In this paper, we present \emph{SemanticNN}, a semantic codec designed to achieve error-resilient \emph{Semantic} \emph{N}eural \emph{N}etwork offloading to achieve compressive transmission correctness at a semantic level for extremely weak devices.
To achieve this, we carefully design the model structure of SemanticNN, including the SQ-based encoder and BER-aware decoder.
We then propose the novel \emph{Feature-augmentation Learning} to enhance the performance of SemanticNN in the presence of transmission errors. 
We further propose the \emph{XAI-based Asymmetry Compensation} method to aid in error-resilient semantic extraction and feature map fine-tuning in the weak encoder.

We conduct extensive evaluations of SemanticNN, demonstrating that it drastically reduces the offloaded features by 56.82-344.83$\times$ and significantly outperforms compressed offloading approaches by 7.11-30.88\% and error resilient works by 1.28-53.42\% with transmission errors in image classification.
Finally, the case study confirms the broad applicability of SemanticNN in real-world scenarios, enabling error-resilient AI in poor communication scenarios.

%
%
%


\section*{Acknowledgments}
We thank all the reviewers for their valuable comments and helpful suggestions.
This work is supported by the National Natural Science Foundation of China under grant no. 62272407 and no. 92582114, the ``Pioneer'' and ``Leading Goose'' R\&D Program of Zhejiang Province under grant No. 2023C01033.

\bibliography{aaai2026}

\appendix

\section{Motivation}
\label{sec:motivation}

\subsection{Compression Rate vs. Error Resilience}
\label{moti:coding}
\begin{figure}[h]
	\centering
	\includegraphics[width=0.98\linewidth]{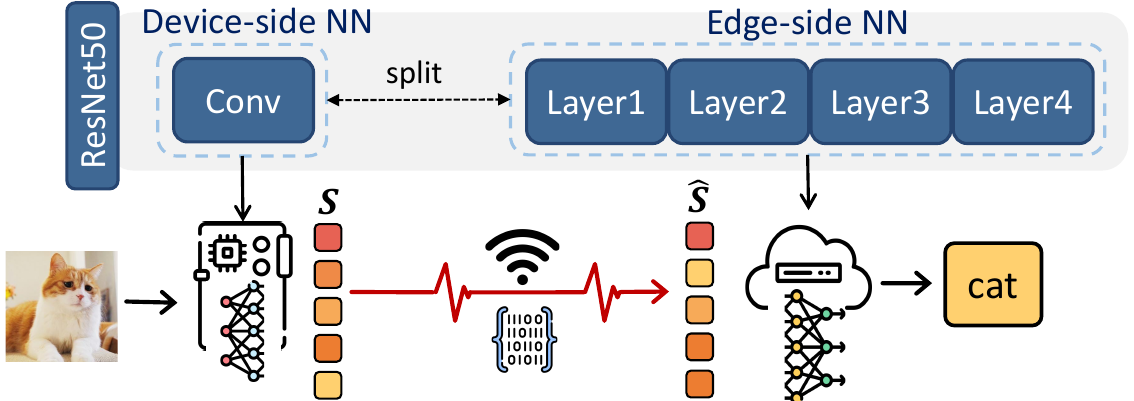} 
	\caption{Motivating example where offloading occurs after the first \emph{Conv} layer of ResNet-50.}
	\label{resnet_split}
\end{figure}

Remarkable efforts~\cite{yao2020deep, laskaridis2020spinn, chen2022context, huang2022real} have been made in minimizing transmission data during device-edge collaboration, leading to compression levels of approximately 0.1\% or even lower.
However, transmission errors are overlooked by existing approaches.
To quantify the impact of transmission errors, we conducted performance measurements of two recent representative approaches, DeepCOD~\cite{yao2020deep} and AgileNN~\cite{huang2022real}, in a wireless link under different BERs.
The validation involved image classification on the ImageNet dataset utilizing ResNet-50~\cite{he2016deep} for split computing as indicated in Figure~\ref{resnet_split}.

\begin{figure}[h]
    \begin{minipage}[t]{0.46\linewidth}
        \centering
        \includegraphics[width=\textwidth]{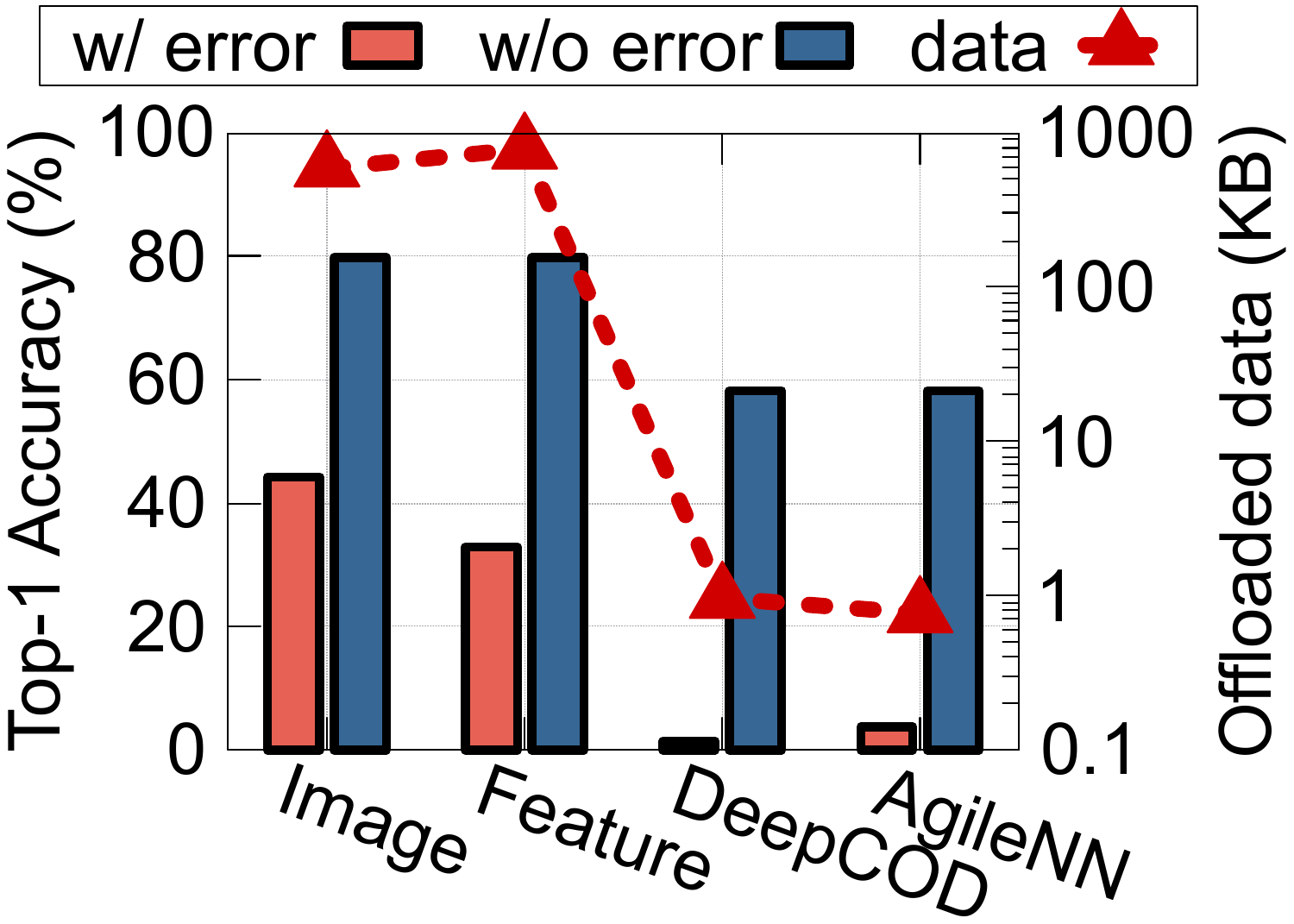}
        \centerline{(a) Performance}
    \end{minipage}%
    \hfill
    \begin{minipage}[t]{0.52\linewidth}
        \centering
        \includegraphics[width=\textwidth]{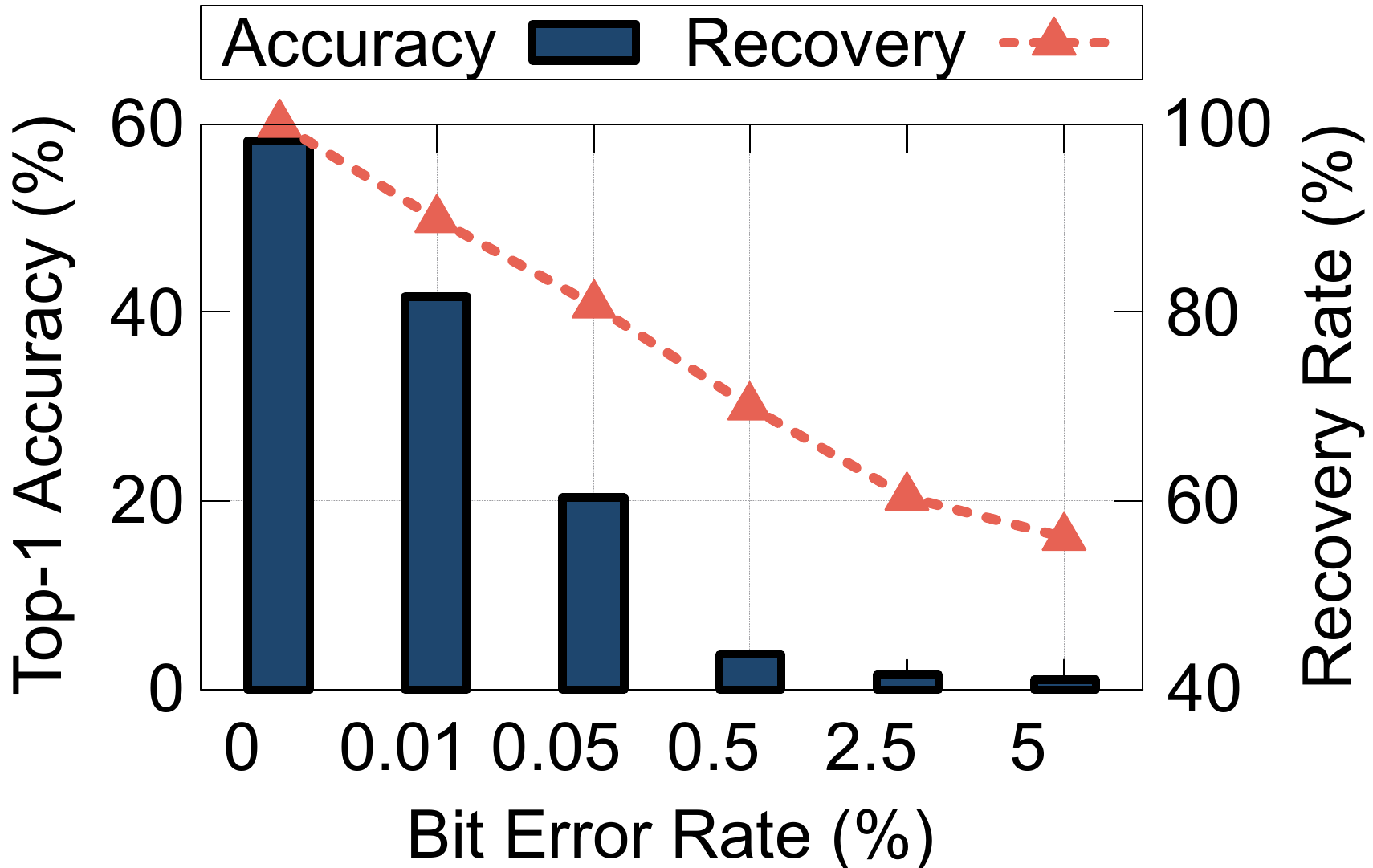}
        \centerline{(b) Recovery rate}
    \end{minipage}
    \caption{(a) Significant accuracy degradation caused by transmission errors.  (b) Even a few bit flips in Huffman-coded data make error correction highly challenging.}
    \label{movitaion_communication}
\end{figure}


Figure~\ref{movitaion_communication} (a) shows the accuracy degradation caused by 0.01-5\% BERs.
Offloading raw images or uncompressed feature maps incurs substantial transmission overhead, leading to an accuracy drop of about 40\%.
While DeepCOD and AgileNN compress the data before offloading, the accuracy degrades significantly. 
This is primarily attributed to their reliance on variable-length coding schemes (e.g., Huffman coding~\cite{huffman1952method}), which are highly sensitive to bit errors during transmission.
Figure~\ref{movitaion_communication} (b) shows the accuracy of DeepCOD~\cite{yao2020deep} using Huffman coding under varying BERs.
The results reveal that even a few bit flips can severely distort the decoded feature maps. 
For instance, at a BER of merely 0.01\%, corresponding to only one bit-flip, approximately 10\% of the features become unrecoverable, resulting in an accuracy drop of around 20\%. 
As the BER increases to 0.5\%, approximately 30\% of features are corrupted, leading to a final accuracy of nearly 5\%. 

Notably, the theoretical recovery rate drops to about 90\% with just one bit flip. 
We optimized the decoding strategy for a reasonable recovery rate, that is, when encountering an invalid Huffman code during decoding, the system skips the current bit and attempts to realign the decoding process.
This allows for partial recovery of subsequent features. 
Thus, the challenge stems from the structure of variable-length codes, which lack explicit boundaries between encoded features. 
A single flipped bit can propagate errors throughout decoding, significantly degrading both feature integrity and final inference accuracy. 
Therefore, reducing transmission overhead while ensuring robustness against channel noise remains a critical challenge.

\subsection{Symmetry vs. Asymmetry}
\label{sec:ays}

The computational limitations of weak devices confine them to executing AI tasks, resulting in an asymmetric structure of AI models on different ends. 
Specifically, the device-side models are significantly smaller in size compared to the edge-side models.
As we will demonstrate, this architectural imbalance introduces notable challenges for collaboration performance.

\begin{figure}[t]
    \begin{minipage}[t]{0.92\linewidth}
        \centering
        \includegraphics[width=\textwidth]{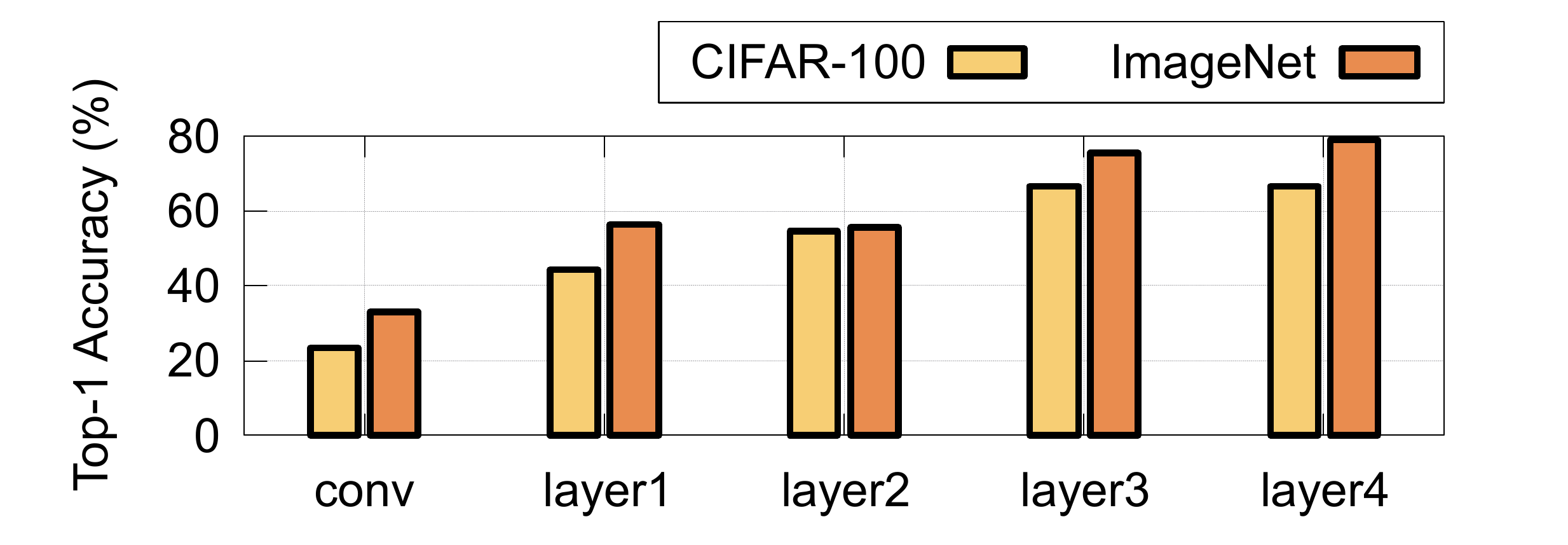}
        \centerline{(a) Top-1 accuracy at different split points}
    \end{minipage}
    \begin{minipage}[t]{1\linewidth}
        \centering
        \includegraphics[width=\textwidth]{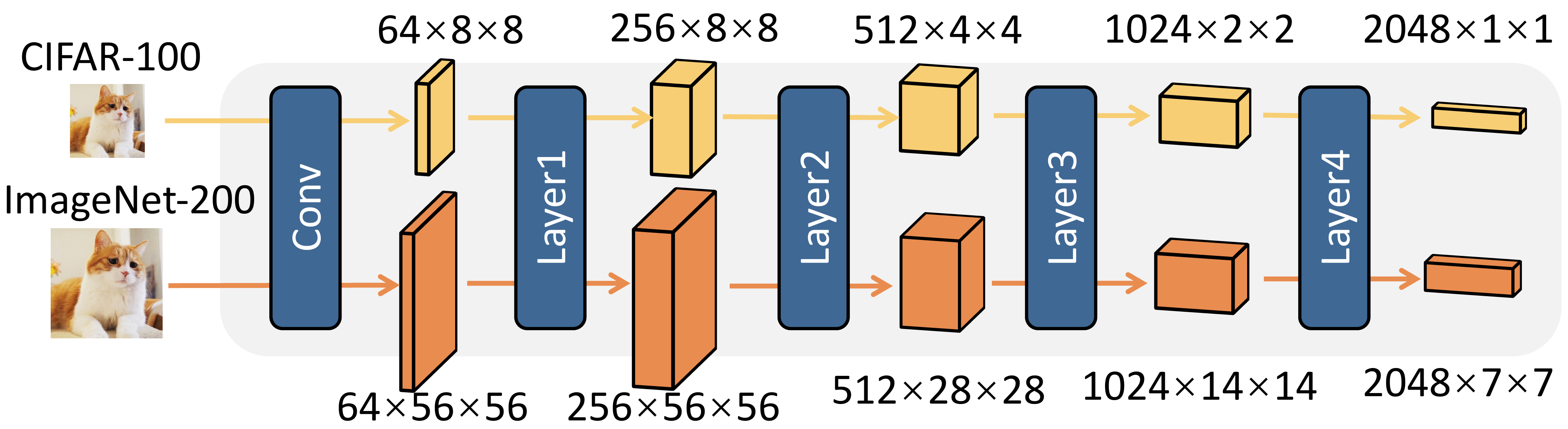}
        \centerline{(b) Amount of data for offloading at different split points}
    \end{minipage}
    \caption{Top-1 accuracy and offloaded data at different split points of ResNet-50 facing BER of 0.01\% to 5\%.}
    \label{movitaion_resnet}
\end{figure}

To investigate the impact of model partitioning, we evaluate ResNet-50 with different split points, as illustrated in Figure~\ref{resnet_split}.
The results, shown in Figure~\ref{movitaion_resnet}, depict the top-1 accuracy across five different split layers on two datasets, i.e., CIFAR-100~\cite{2009Learning} and ImageNet-200.
We observe that placing more layers on the device side generally improves inference accuracy on both datasets.
The highest accuracy is achieved when splitting after layer 4, where the amount of offloaded data remains relatively small, as shown in Figure \ref{movitaion_resnet} (b). 
Although this split configuration appears optimal in terms of accuracy and communication overhead, it is not feasible in practical device-edge collaboration scenarios. 
The limited resources of weak devices, such as the STM32 platform, only allow for a minimal device-side NN, typically consisting of just the initial convolutional layer of ResNet-50. 
Consequently, the structural asymmetry between the device-side and edge-side components severely degrades the overall inference performance.

\begin{figure}[t]
\centering
\includegraphics[scale=0.17]{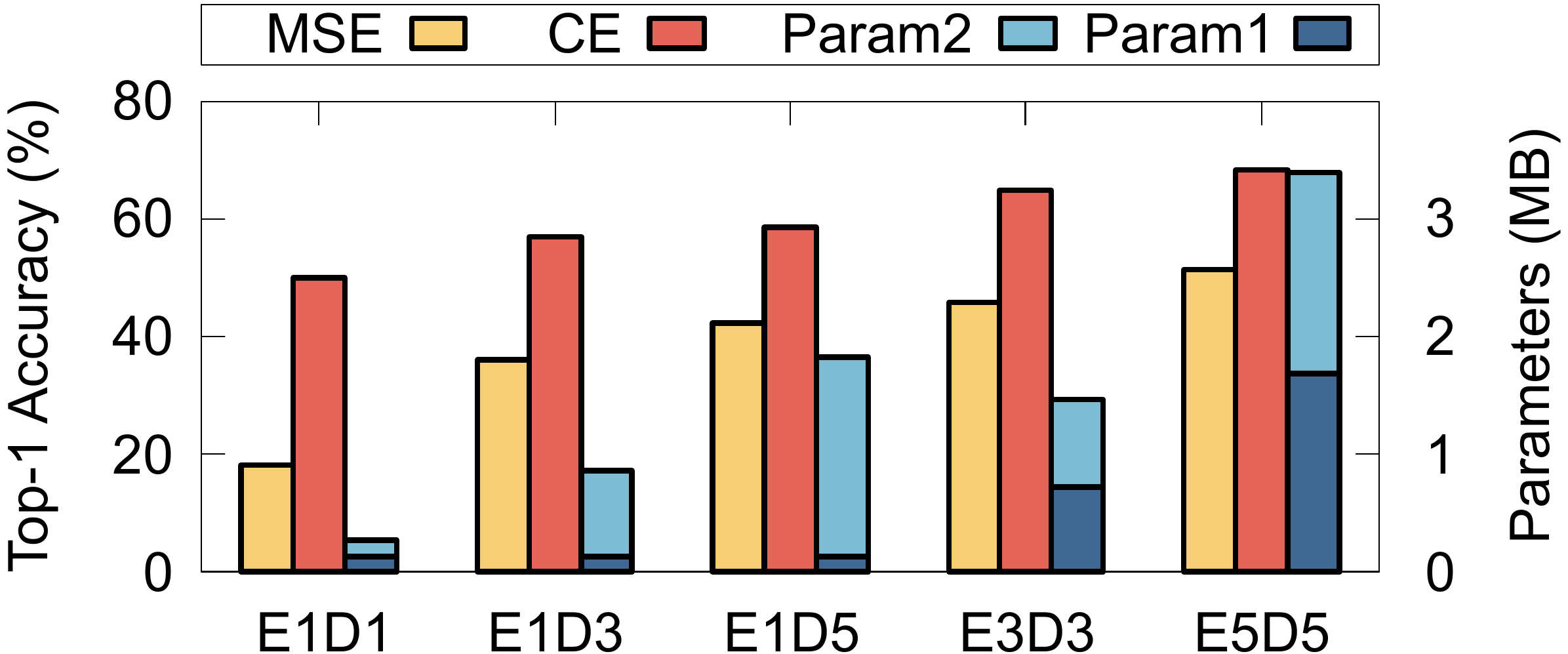}
\caption{Performance and parameters of symmetric and asymmetric ADJSCC for different layer numbers.}
\label{adjscc}
\end{figure}

{Moreover, we investigate the impact of encoder-decoder asymmetry, i.e., the disparity in model size between the encoder and decoder. 
We adopt ADJSCC~\cite{xu2021wireless} as a baseline approach and construct five different encoder-decoder configurations, denoted as E$m$D$n$, where m and n represent the number of layers in the encoder and decoder, respectively. 
Figure~\ref{adjscc} expected that increasing the number of layers generally improves inference accuracy. }
More importantly, symmetric configurations (e.g., E3D3) achieve higher accuracy compared to asymmetric ones (e.g., E1D3 and E1D5).
These observations highlight that due to the resource constraints of weak devices, deploying a large encoder is infeasible, inevitably leading to an asymmetric architecture with a small encoder and a powerful decoder. 
To address this issue, we introduce a novel XAI-based asymmetry compensation technique that enhances feature extraction on the encoder side without introducing additional computational overhead.

\subsection{Pixel-level feature offloading}
\label{sec:mot_pix}
\begin{figure}[h]
\centering
\includegraphics[scale=0.33]{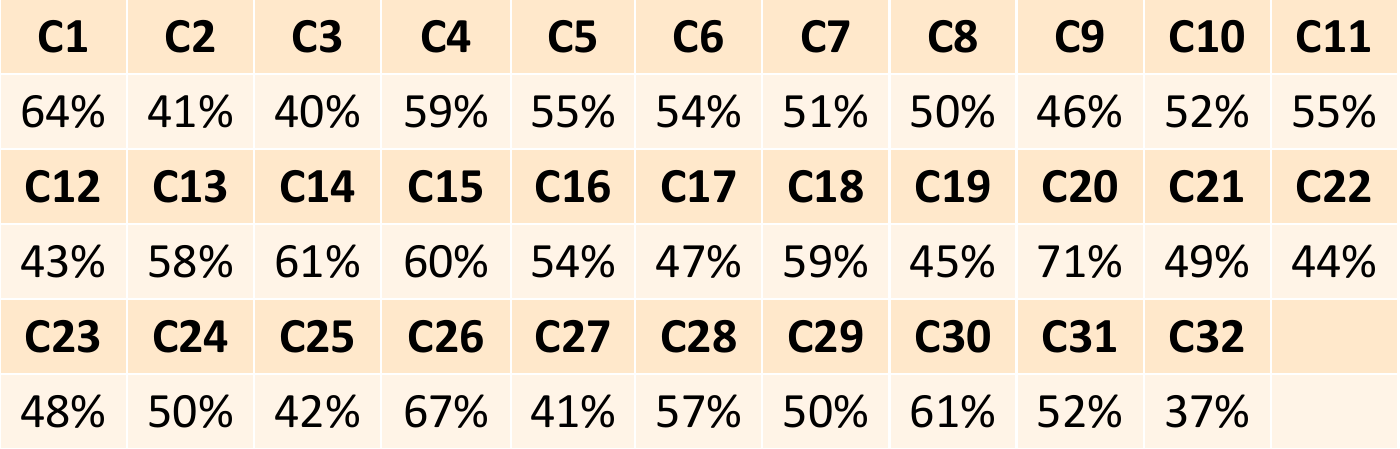}
\caption{Proportion of important pixels in 32 different channels for all offloaded intermediate features.}
\label{channel_level}
\end{figure}

Our experimental findings in Figure~\ref{channel_level} reveal that approximately half of the pixels within a feature play a pivotal role with positive values in its learning process.

\section{Evaluation Details}
\label{sec:evalution_d}
\subsection{Implementation}
\label{sec:imp_d}

We train splitting task models using the SGD optimizer with a learning rate of 0.1 for 300 epochs. 
The codec in SemanticNN is trained using the Adam optimizer with a learning rate of 0.005, encompassing 30 epochs for larger-scale datasets (comprising large-sized images) and 100 epochs for smaller-scale datasets.
In addition, we add a learning rate scheduler in SemanticNN to reduce the learning rate by 0.2 at the midpoint and later stages of training.

After training, the device-side NN of the original model and the encoder in SemanticNN were deployed on the device STM32H750XBH6. 
It is equipped with an ARM 32-bit Cortex-M7 CPU at 400MHz, 1069KB SRAM, 128KB FLASH, and 500MB NAND FLASH storage.
The PyTorch format model files were converted to ONNX format files and subsequently integrated into the program using CubeIDE.
Meanwhile, the decoder and the edge-side NN were deployed on the edge server, utilizing an NVIDIA JETSON NX server.
\subsection{Datasets} 

The datasets used for training SemanticNN and evaluating semantic offloading include the following two categories, image classification (ImageNet-200, CIFAR-100~\cite{2009Learning}, and RSI-CB256~\cite{li2017rsi}) and object detection (COCO~\cite{lin2014microsoft}, PASCAL-VOC2007, and NWPU VHR-10~\cite{cheng2016survey}).

Datasets for image classification:
\begin{itemize}
	\item \emph{ImageNet-200}, a subset of ImageNet, comprises 200 classes of 224x224 RGB images, with 50,000 for training and 10,000 for validation/testing each. 

	\item \emph{CIFAR-100} dataset contains 32×32 RGB images, composed of 50,000 training and 10,000 testing images, corresponding to 100 classes. 
	
	\item \emph{RSI-CB256} is a satellite dataset consisting of four categories from various sensors and Google Maps.
\end{itemize}
Datasets for object detection:
\begin{itemize}
	\item \emph{COCO} dataset comprises 328,000 images and a staggering 2,500,000 labels including 91 target classes. 
		
	\item \emph{PASCAL-VOC2007} comprises 20 different categories of objects, totaling 9,963 images, and includes a total of 24,640 labeled target items.
		
	\item \emph{VHR-10} is a detailed 10-level geo-remote sensing dataset featuring 800 images across 10 classes, including 650 with specific targets and 150 background scenes.
\end{itemize}

\subsection{Baselines} 
We primarily compare our work in terms of data compression and error resilience with the following baselines.
We utilize pre-trained ResNet-50~\cite{he2016deep}, MobileNetV2~\cite{sandler2018mobilenetv2}, and YOLOv5~\cite{YOLOv5} as original models in split computing. 
Since the efficient coding used in existing approaches is variable-length and highly unstable with transmission errors (e.g., lower than 3\% top-1 accuracy for DeepCOD~\cite{yao2020deep} with errors), we made small improvements (i.e., binary fixed-length coding) for existing approaches, DeepCOD and AgileNN, to obtain a fair and comprehensive comparison.
\begin{itemize}
	\item \emph{Image} is directly offloaded without split computing.
	
	\item \emph{Feature} is offloaded without any processing.
	\item \emph{DeepCOD*} replaces the Huffman coding in DeepCOD~\cite{yao2020deep} with binary coding.
	
	\item \emph{AgileNN*} also replaces coding in AgileNN~\cite{huang2022real} with binary coding and tests in an offloading scenario at a 20\% split ratio.
	
	\item \emph{A-ADJSCC} employs an asymmetric structure modified for ADJSCC~\cite{xu2021wireless} to accommodate the asymmetric computational power in split computing.
	
	\item \emph{NeuroMessenger}~\cite{wang2022neuromessenger} adaptively adjusts the retransmission mechanism based on user requirements to balance communication overhead and task performance.
	
	\item \emph{DeepSC}~\cite{xie2021deep} employs a Transformer structure to extract semantic information for text transmission.
\end{itemize}

\section{Overhead}
\label{sec:overhead}
We also evaluate the memory consumption of device-side NN and the encoder of SemanticNN on STM32H750XBH6. 
For MobileNetV2 on CIFAR-100 and RSI-CB256, the total parameters are less than 100KB, and the memory footprint in RAM is less than 400KB. Due to the small size of the offloaded features, the communication time becomes negligible.
\begin{table}[h]
    \centering
    \begin{tabular}{ccccc}
        \toprule
         \textbf{Image} & \textbf{Model} & \textbf{Params} & \textbf{Memory} \\
         \midrule
         \midrule
         \multirow{2}{*}{3*64*64} & MobileNetV2 & 2.62KB & 193.42KB\\		
         & SemanticNN & 52.26KB & 196.13KB\\	
         \midrule	
         \multirow{2}{*}{3*32*32} & MobileNetV2 & 2.62KB  & 49.42KB\\		
          & SemanticNN & 52.26KB & 52.13KB\\		
        \bottomrule
    \end{tabular}
    \caption{Memory consumption on STM32H750.}
    \label{overhead}
\end{table}

\section{Impact of Feature-augmentation Learning}
\label{two_stage}
To maintain the task performance while facing communication transmission errors, we propose a progressive Feature-augmentation Learning progress for SemanticNN.

\begin{figure}[h]
    \begin{minipage}[t]{0.5\linewidth}
        \centering
        \includegraphics[width=\textwidth]{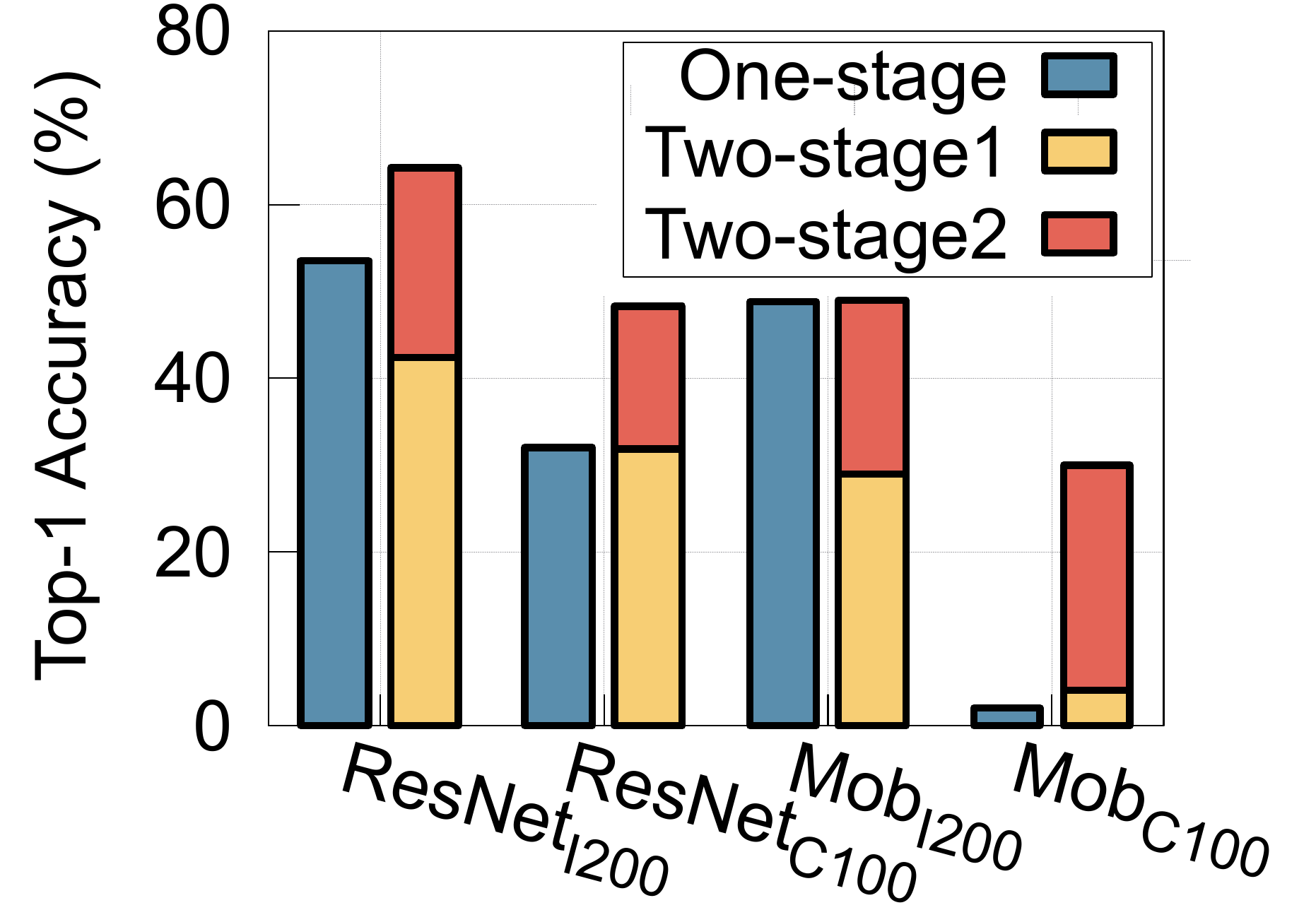}
        \centerline{(a) One-stage vs. Two-stage}
    \end{minipage}%
    \begin{minipage}[t]{0.5\linewidth}
        \centering
        \includegraphics[width=\textwidth]{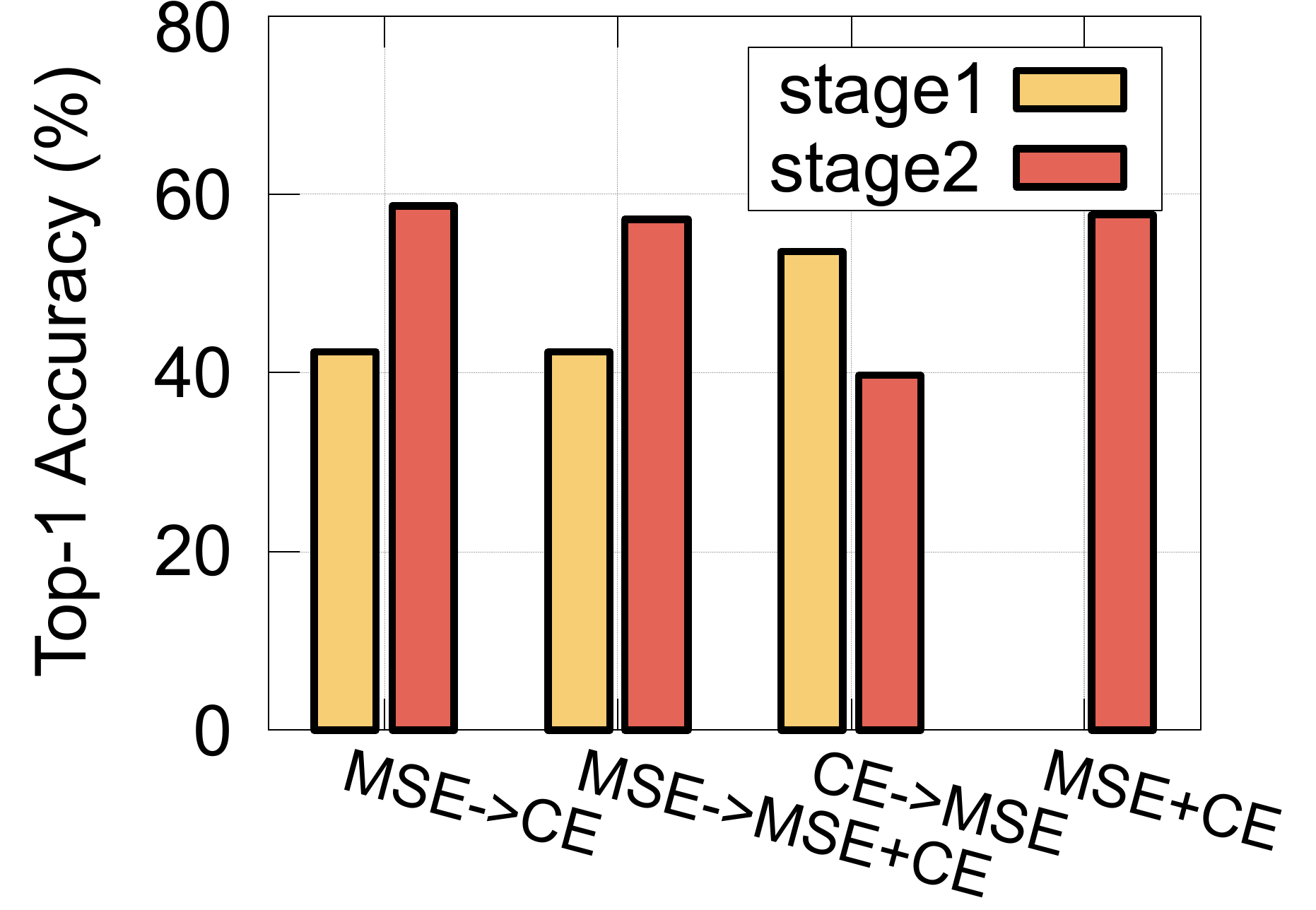}
        \centerline{(b) Stage1 \& Stage2}
    \end{minipage}
    \caption{(a) The two-stage Feature-augmentation Learning benefits for task performance. (b) The design of the corresponding loss function in different stages is also reasonable.}
    \label{two_stage_accuracy}
\end{figure}

Figure~\ref{two_stage_accuracy} (a) demonstrates that this workflow yields superior performance compared to a one-stage training that solely focuses on semantic objectives (i.e., Task-oriented Semantic-level Training).
Furthermore, we also validate the respective efficacy and its loss function of the two-stage model training, whether it is task performance (CE) or autoencoder performance (MSE).
Figure~\ref{two_stage_accuracy} (b) illustrates that prioritizing the denoising autoencoder performance in the first stage and subsequently focusing on task performance on the semantic level in the second stage is a reasonable approach.
Simultaneously considering both in the second stage also yields comparable results when the offloading features extracted by device-side NN are inherently error-resistant.

\section{Real-world cases}
\label{sec:case}
\begin{figure}[h]
\centering
    \includegraphics[scale=0.1]{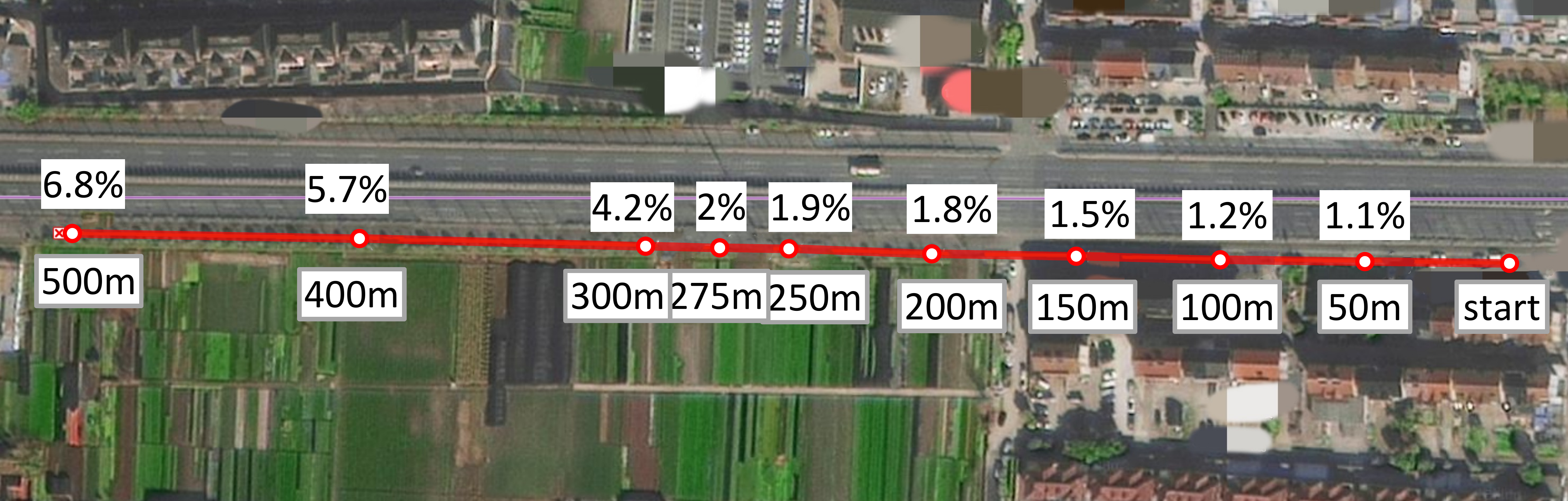}
    \caption{LoRa's BER vs. distance, featuring 812.5 kHz bandwidth \emph{without} any error-control network codec.}
    \label{lora}
\end{figure}

\begin{figure}[h]
\centering
    \includegraphics[scale=0.23]{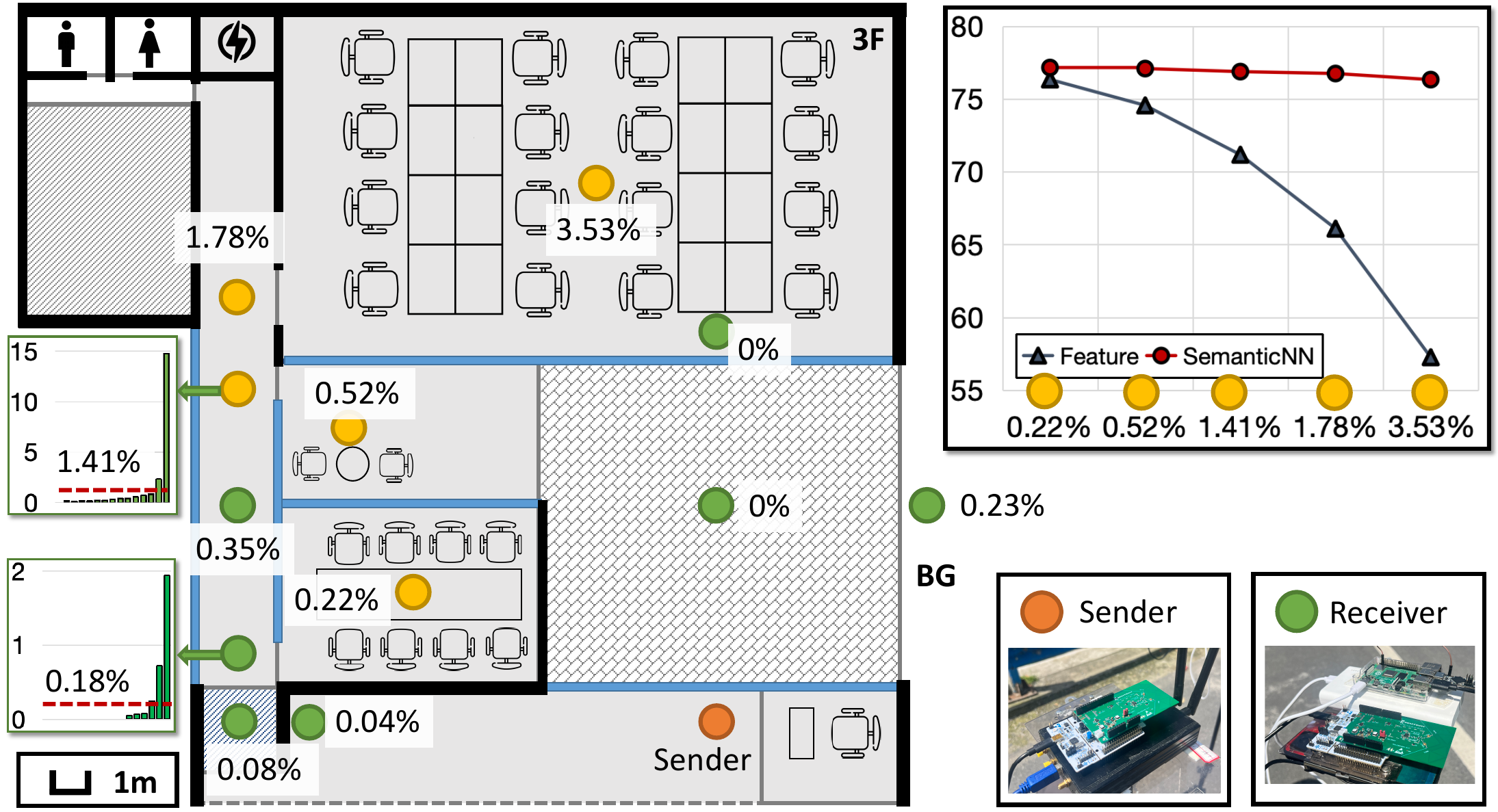}
    \caption{SemanticNN adapts to various BERs across different positions in real-world indoor Wi-Fi scenarios and maintains superior and stable performance of various applications.}
    \label{fig:case_study}
\end{figure}

To elucidate the relationship between simulated BERs and real-world conditions and assess the applicability of SemanticNN, we conducted case studies for outdoor LoRa and indoor Wi-Fi communications, as depicted in Figure~\ref{lora} and ~\ref{fig:case_study}. 

The percentage value in Figure~\ref{lora} and ~\ref{fig:case_study} presents the mean BERs observed over time at specific locations.

The actual BER varies 1.1-6.8\% with varying distance in the outdoor LoRa link and
0-3.53\% with varying levels of occlusion, distance, and altitude difference in indoor Wi-Fi link. 
The right part of Figure~\ref{fig:case_study} shows the classification accuracy of ResNet-50 on ImageNet-200 in relation to communication BER. 
SemanticNN demonstrates enhanced tolerance to BER and maintains superior and stable task performance when compared to methods that rely solely on bit-level error control for the direct offloading of intermediate features.

Moreover, it is imperative to highlight that BER fluctuates dynamically even at a fixed location, underscoring the dynamic nature of communication environments. 
This observation further substantiates the adaptability of SemanticNN to varying BER conditions within a large range of applications.
In conclusion, SemanticNN can further break through the performance limitations (communication overhead, distance, etc.) based on bit-level optimization in real-world scenarios.

\section{Parameter Study}
\label{param_study}

\begin{figure}[h]
    \begin{minipage}[h]{0.5\linewidth}
        \centering
        \includegraphics[width=\textwidth]{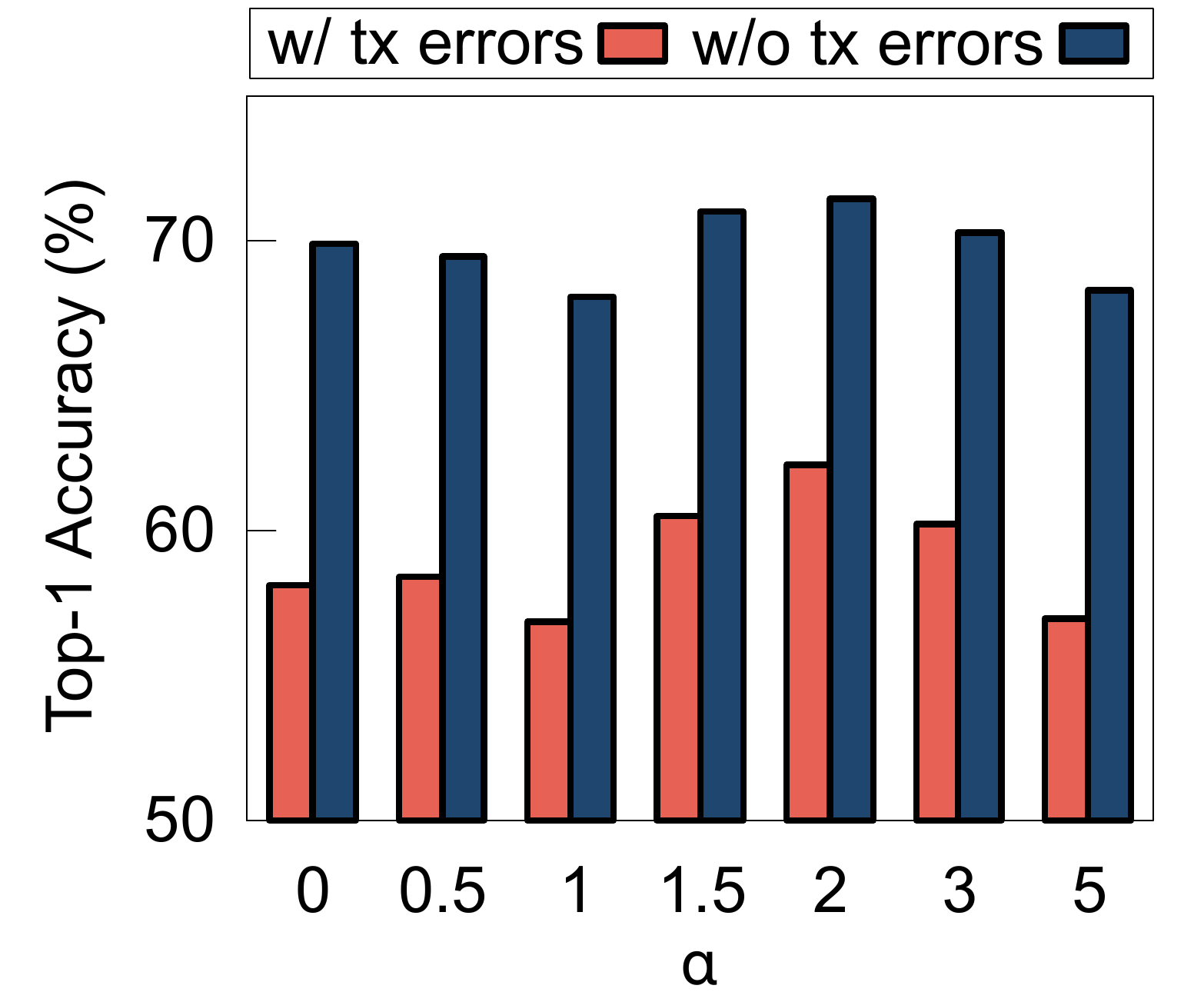}
        \centerline{(a) Top-1 accuracy}
    \end{minipage}%
    \begin{minipage}[h]{0.5\linewidth}
        \centering
        \includegraphics[width=\textwidth]{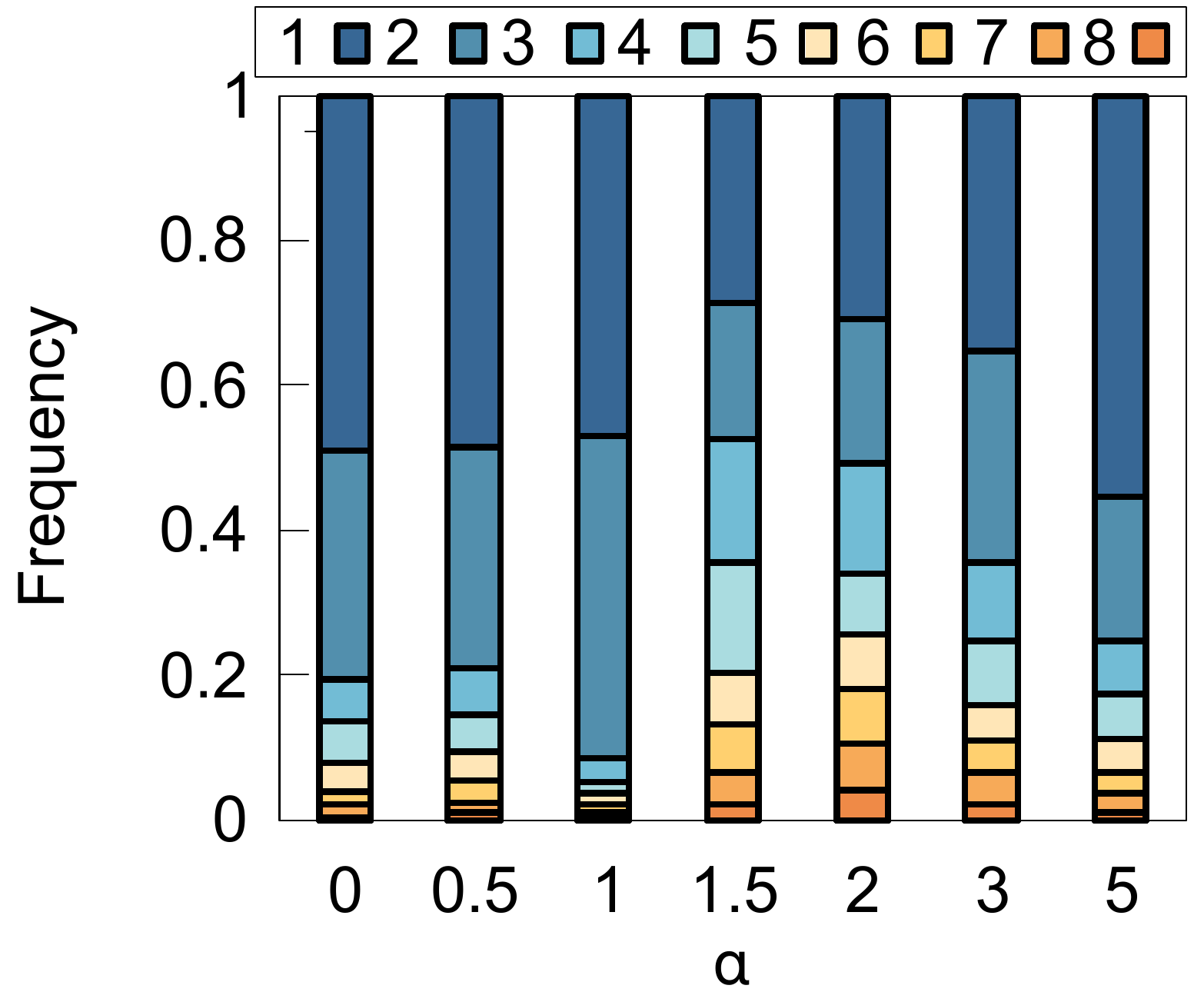}
        \centerline{(b) Quantization centers}
    \end{minipage}
    \caption{When $\alpha$ is $2$, the offloaded data can be quantified perfectly, and SemanticNN achieves better task performance with transmission errors.}
    \label{low_entropy}
\end{figure}

\textbf{Impact of $\alpha$ for $\mathcal{L}_{Q}$}.
$\alpha$ represents the importance of soft quantization in the final loss function.
Figure~\ref{low_entropy} (b) illustrates that the frequency of quantization centers is the most balanced (i.e., highest entropy) when $\alpha$ is set to $2$. 
Additionally, relatively high accuracy is also achieved when $\alpha$ is $2$, shown in Figure~\ref{low_entropy} (a).
Therefore, the design and implementation of the balanced-learning quantization method indeed contribute to the final performance gain of SemanticNN.

\begin{figure}[h]
    \begin{minipage}[h]{0.5\linewidth}
        \centering
        \includegraphics[width=\textwidth]{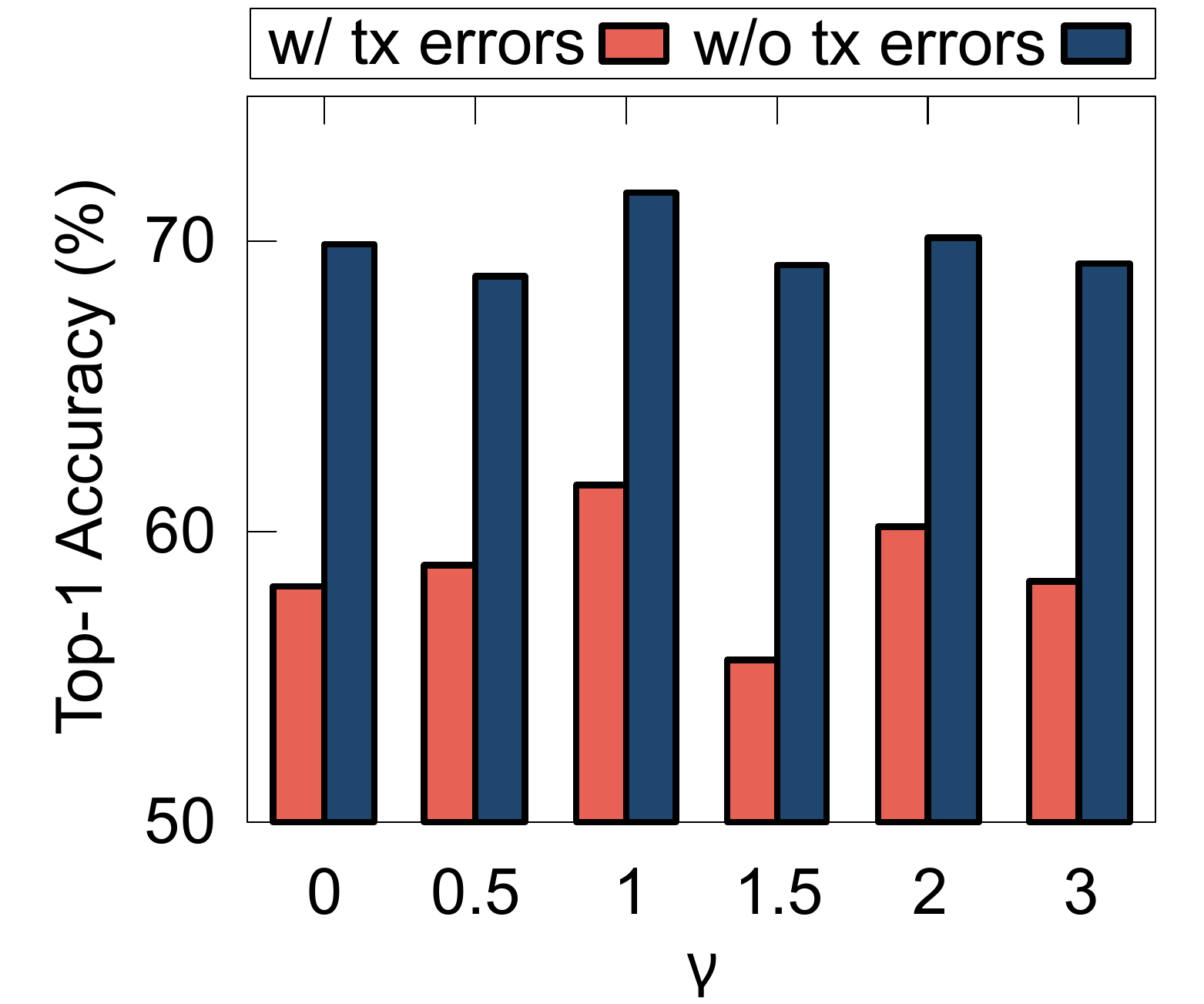}
        \centerline{(a) Top-1 accuracy}
    \end{minipage}%
    \begin{minipage}[h]{0.5\linewidth}
        \centering
        \includegraphics[width=\textwidth]{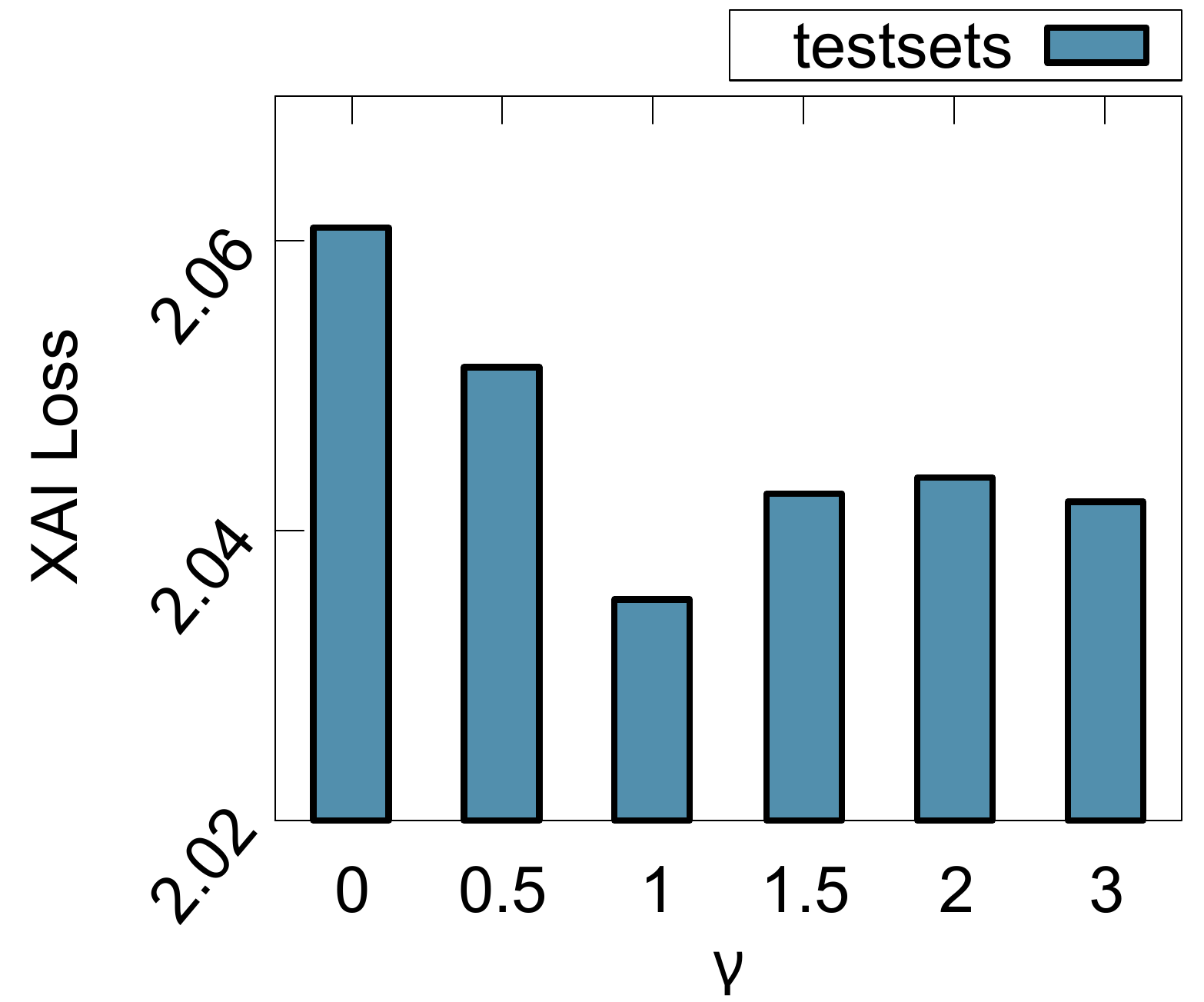}
        \centerline{(b) XAI loss}
    \end{minipage}
    \caption{When $\gamma$ is $1$, the \emph{XAILoss} of offloaded data is low, showing that the capability of the encoder in SemanticNN is enhanced and achieves better performance.}
    \label{xai_results}
\end{figure}

\textbf{Impact of $\gamma$ for $\mathcal{L}_{XAI}$}.
$\gamma$ represents the importance of the offloading features in the final loss function.
Figure~\ref{xai_results} (b) shows that the lowest loss among all configurations is achieved when $\gamma$ is set to be $1$.
Accordingly, Figure~\ref{xai_results} (a) also shows that the relatively satisfactory task performance is achieved when $\gamma$ is $1$ with or without transmission errors, which does make the encoder transmit the offloaded data containing more important features.

\end{document}